\documentclass[11pt]{article}
\usepackage[T1]{fontenc}
\usepackage{fancyhdr}  
\usepackage[a4paper,margin=1in]{geometry}
\usepackage[utf8]{inputenc}
\usepackage{amsmath,amssymb,amsfonts}
\usepackage{graphicx}
\usepackage{microtype}
\usepackage{xcolor}
\usepackage{hyperref}

\usepackage{caption}
\usepackage{subcaption}
\usepackage{tabularx}
\usepackage{enumitem}
\usepackage{natbib}
\usepackage{setspace}
\usepackage{booktabs}
\usepackage{siunitx}
\usepackage{threeparttable}
\usepackage{adjustbox}
\usepackage{float}
\usepackage{array}
\usepackage[section]{placeins}
\usepackage{chngcntr}
\usepackage{indentfirst}

\newcommand{\cell}[2]{\num{#1}\,(\num{#2})}  % mean (std)
\newcommand{\shead}[1]{\textbf{#1}}
\newcommand{\theadsm}[1]{\textbf{\small #1}}
\newcommand{\unifiedtabsetup}{%
  \footnotesize
  \setlength{\tabcolsep}{3.8pt}%
  \renewcommand{\arraystretch}{1.16}%
}

\hypersetup{
  colorlinks=true,
  linkcolor=blue!60!black,
  citecolor=blue!60!black,
  urlcolor=blue!60!black
}

\newcommand{\Lag}{T_{\text{lag}}}

\setlist{nosep,leftmargin=*}

\title{Spatiotemporal Satellite Image Downscaling with Transfer Encoders and Autoregressive Generative Models}

\author{
  Yang Xiang\textsuperscript{1},
  Jingwen Zhong\textsuperscript{1,+},
  Yige Yan\textsuperscript{1,+},
  Petros Koutrakis\textsuperscript{2},
  Eric Garshick\textsuperscript{3,4},\\
  Meredith Franklin\textsuperscript{1,*}
  \\
  \\
  \small \textsuperscript{1}Department of Statistical Sciences, University of Toronto \\
  \small \textsuperscript{2}Harvard T.H. Chan School of Public Health \\
  \small \textsuperscript{3}Harvard Medical School \\
  \small \textsuperscript{4}VA Healthcare System Boston, U.S. Department of Veterans Affairs \\
  \small \textsuperscript{+}These authors contributed equally. \\
  \small \textsuperscript{*}Corresponding author: \texttt{meredith.franklin@utoronto.ca}
}

\date{}

\begin{document}
\maketitle

% ============================ ABSTRACT (placeholder) ============================
\begin{abstract}
We present a transfer-learning generative downscaling framework to reconstruct fine-resolution satellite images from coarse inputs. Our approach combines a lightweight temporal U-Net transfer encoder with a diffusion-based generative model to produce daily fine resolution downscaled images over a 20-year period. The simpler U-Net is first pretrained on the long time series of coarse resolution data to learn spatiotemporal representations; its encoder is then frozen and transferred to a larger downscaling model as physically meaningful latent features. Our application uses NASA’s MERRA-2 reanalysis as the low-resolution source domain ($\sim$ 50 km resolution) and the GEOS-5 Nature Run (G5NR) as the high-resolution target ($\sim$ 7 km resolution). Our study area included a large area in Asia, which was made computationally tractable by splitting into two subregions and four seasons. We conducted domain similarity analysis using Wasserstein distances confirmed minimal distributional shift between MERRA-2 and G5NR, validating the safety of parameter-frozen transfer.
Across seasonal–regional splits, our model achieved excellent performance (R$^2$ = 0.65 - 0.94), outperforming comparison models including deterministic U-Nets, variational autoencoders, and prior transfer-learning baselines. Out-of-data evaluations using  semivariograms, ACF/PACF, and lag-based RMSE/R$^2$ demonstrated that the predicted downscaled images preserved physically consistent spatial variability and temporal autocorrelation, enabling stable autoregressive reconstruction beyond the G5NR record. These results show that transfer-enhanced diffusion models provide a robust and physically coherent solution for downscaling a long time series of coarse resolution images with limited training periods. This advancement has significant implications for improving environmental exposure assessment and long-term environmental monitoring.
\end{abstract}

% ============================ 1. INTRODUCTION (placeholder) =====================
\section{Introduction}\label{sec:intro}

Deep learning has achieved breakthroughs across vision, speech and recommender systems \citep{10.1145/3065386, article, 10.1145/3285029}, and has increasingly advanced applications in remote sensing of earth observations \citep{YUAN2020111716, Li2020_dl}. Recent progress shows that neural networks can learn complex spatial and temporal structure from data \citep{li2023}, making them well-suited to transform coarse-resolution inputs into high-resolution predictions. This capability is central to spatial downscaling, a task that lies at the intersection of machine learning and environmental modeling. While traditional statistical downscaling methods have long served to bridge Global Climate Model (GCM) outputs and local observations \citep{wilby_statistical_1998}, its limited ability to capture nonlinear multiscale dependencies motivates machine learning alternatives. Deep learning frameworks explicitly model such dependent variables to reconstruct fine-resolution geophysical states from coarse inputs \citep{yuan_deep_2020, bano-medina_configuration_2020}. In particular, convolutional neural networks (CNNs) have shown strong potential for modeling spatial variability, and ensemble/stacked approaches further improve fidelity over diverse regions \citep{bano-medina_configuration_2020, zhang_spatial_2024}.

Beyond deterministic predictors, generative models have emerged as powerful tools for downscaling: they can better capture fine-scale spatial variability relative to conventional convolutional neural network baselines and naturally support probabilistic uncertainty quantification \citep{watt_generative_2024}. Diffusion-based methods have been used to reconstruct long records with local detail while drawing stochastic samples to characterize uncertainty \citep{ling_diffusion_2024}, and conditional generative adversarial networks (GANs) have produced high-resolution geospatial fields from coarse climate drivers, including behavior under extremes \citep{li_generative_2024}. In parallel, transfer learning can leverage data-rich regions or variables to improve performance where data are sparse; for aerosols specifically transfer-augmented networks have generated fine-scale fields from coarser inputs while reducing the need for extensive retraining \citep{Wang2022}. Together, these advances point toward probabilistic, transfer-aware deep learning frameworks for high-resolution climate data production.

Satellite earth observations illustrate both the need and opportunity for such methods. Deep learning–based downscaling of satellite data has yielded high-resolution aerosol optical depth (AOD) and particulate matter estimates from coarse inputs \citep{Li2020_dl, Wang2022}. Aerosols play a critical role in global climate, air quality, and health. AOD, which is commonly characterized using satellite or ground-based remote sensing products, is a unitless vertically integrated measure of the atmospheric extinction caused by aerosols. It quantifies the total amount of sunlight in a column of the atmosphere that is prevented from reaching the ground due to the combined effects of scattering and absorption by aerosol particles. In this study, we focus mainly on dust AOD visibility-relevant quantity with documented health impacts \citep{atmos7120158}. High-quality fine-scale aerosol information is crucial for exposure assessment, particularly where natural and anthropogenic sources co-occur. Yet ground-based networks remain critically sparse in space and time. Satellite products have been instrumental in filling the gaps where ground-based networks are lacking. Satellite observations coupled with climate and chemical transport models have been synthesized in products such as NASA's Modern-Era Retrospective analysis for Research and Applications, Version 2 (MERRA-2).

We present a two-stage transfer learning downscaling framework for enhancing the resolution of gridded satellite products, using dust extinction AOD as an example. The first stage employs a small U-Net encoder, pretrained on the coarser resolution data to learn robust spatiotemporal representations of atmospheric dynamics from time series data. The encoder weights are then frozen to preserve the learned long-term spatiotemporal associations and used as a feature extractor for the second-stage large model. In the larger downscaling model, the encoded features are rescaled to align with a high-resolution input grid and then incorporated into a Denoising Diffusion Probabilistic Model (DDPM), which predicts high-resolution output maps. The DDPM integrates multiple inputs, including the transfer features from the small model, current-day coarse resolution data, and spatial and temporal indices. It learns via a smoothly varying noise process and constrains its final predictions to remain within realistic ranges using a sigmoid activation. For comparison, we also implement other deterministic and probabilistic models, including a supervised U-Net and a FiLM-conditioned VAE, following the same transfer pipeline. %Models are trained separately for each season (DJF, MAM, JJA, SON) \citep{w17101475} and region (i) Afghanistan–Kyrgyzstan and (ii) the Arabian Peninsula–Mesopotamian basin, which includes United Arab Emirates, Iraq, Kuwait, Qatar, Saudi Arabia, and Djibouti. 
Our results demonstrate that incorporating transferred features from the coarser resolution data improves spatial fidelity at finer resolutions, and most importantly, enable downscaling beyond the training window when fine resolution data are not available.

% ============================ 2. Method =========================
\section{Methods}\label{sec:Method}

\subsection{Data Preparation}\label{sec:data preparation}

\paragraph{Satellite Re-Analysis and Models: MERRA\mbox{-}2 and G5NR}
The Modern-Era Retrospective analysis for Research and Applications, Version~2 (MERRA\mbox{-}2) is a NASA Global Modeling and Assimilation Office (GMAO) reanalysis of space-based aerosol observations and their interactions in the climate system \cite{gelaro_modern-era_2017}. It provides global 50 km\(0.5^{\circ}\!\times\!0.625^{\circ}\) gridded estimates of different types of aerosols and climate variables at an hourly time resolution. In this study, we present the downscaling framework using MERRA-2's dust extinction aerosol optical depth (AOD) at 550\,nm, which is a measure of the total column extinction of light specifically due to mineral dust aerosols in the atmosphere. Of the satellite aerosol properties, extinction AOD is the easiest to retrieve, so it was chosen for its reliability \cite{bakatsoula2023} and relevance to the study region (described below).

We spatially clipped the hourly images to two regions: Afghanistan-Kyrgyzstan (Area 0) and Southwest Asia (Area 1 which included the United Arab Emirates, Iraq, Kuwait, Qatar, Saudi Arabia, and Djibouti), and calculated daily averages over the study period 2000-01-01 to 2024-12-31 to produce daily images. The geographic splits were chosen to reflect different climate and terrain conditions. The data were stored as three-dimensional arrays with axes \(\text{day}\times\text{latitude}\times\text{longitude}\) and an associated dust extinction value (unitless).

To provide a physically consistent high-resolution reference field for downscaling, we use the GEOS-5 Nature Run (G5NR), a free-running global simulation produced with the same GEOS modeling system. Unlike MERRA-2 (a reanalysis with observation assimilation), G5NR is an unconstrained "nature run" that resolves mesoscale dynamics and includes fully interactive aerosols. It is provided at \(0.0625^{\circ}\) ($\sim$ 7\,km) grid with nominal 30-minute output. We aggregate the 30-minute 550 nm dust-extinction AOD to daily UTC means and restrict the record to the documented G5NR window (June 2005 through May 2007). After regional clipping, the storage layout matches MERRA\mbox{-}2: \(\text{day}\times\text{latitude}\times\text{longitude}\), with each daily image giving pixelwise dust–extinction intensity. For downscaling, we use the MERRA-2-G5NR temporal intersection (June 2005–May 2007) and regrid MERRA-2 to the G5NR grid (described below).

Elevation is used as an additional spatial predictor, and is taken from the Global Multi-resolution Terrain Elevation Data 2010 (GMTED2010) at 30-arc-second (\(\approx\)1\,km) resolution. It is constant over time and is stored as a two-dimensional latitude-longitude array.

Additional pre-processing included log-transforming dust extinction AOD for numerical stabilization because it is skewed. Because MERRA\mbox{-}2 ($0.5^{\circ}\!\times\!0.625^{\circ}$), G5NR ($0.0625^{\circ}$), and elevation ($\approx$30-arc-second) have different native resolutions, we first cropped each dataset to a common rectangular analysis window that tightly contained the countries of interest. Daily MERRA\mbox{-}2 images were then interpolated to the centers of G5NR pixels using bicubic interpolation, and elevation was aggregated to the G5NR resolution by averaging all native 30-arc-second samples within each pixel. After this step, daily MERRA\mbox{-}2 images, elevation, and the latitude/longitude coordinate arrays were pixel-aligned one-to-one with the G5NR images inside the analysis window.

Each region was further split into four seasons based on standard climatological definitions: December–February (DJF), March–May (MAM), June–August (JJA), and September–November (SON). This follows the convention of the World Meteorological Organization and recent seasonal dust studies in the Middle East \citep{w17101475}. For each season, we added a 45-day buffer period before the start of the season. The buffer ensures that for each target day, the model has a full sequence of recent days as input. While buffer days were not used as prediction targets, they provide valid historical context for constructing the lag sequence.

Prediction targets were limited to days within each season window. Within each Area×Season split we split days in temporal order: the first \(80\%\) for training, the next \(10\%\) for validation, and the final \(10\%\) for testing. There was no mixing across days. All standardization parameters were estimated on training days only by pooling all training
pixels across days and variables, and the same parameters were applied unchanged to validation,
test, and out-of-data periods. The season index remained an integer label and was not standardized.

We trained on mini-batches of \(16{\times}16\) patches cut from each daily image at the G5NR resolution. A sliding window with stride \(8\) pixels in both directions created overlapping patches and increased the effective sample size. For a given target day, we first assembled that day’s inputs, then extracted all \(16{\times}16\) patches with stride \(8\), discarded patches that fell outside the valid spatial window, and batched the rest; batch size is fixed per run to fit GPU memory. To prevent data leakage from spatial cropping, the batching pipeline enforced day-level grouping: all patches cut from one day appeared only in training or only in validation within an epoch, never in both. Together with the temporal split and train-only scaling, this prevented leakage across areas and seasons and avoided using future information to normalize past data.

\subsection{Modeling}
\label{sec:modeling}

% Overview of our model: include introduction of datasets and input and output of our model

\begin{figure}[H]
    \centering
    \includegraphics[width=0.98\linewidth]{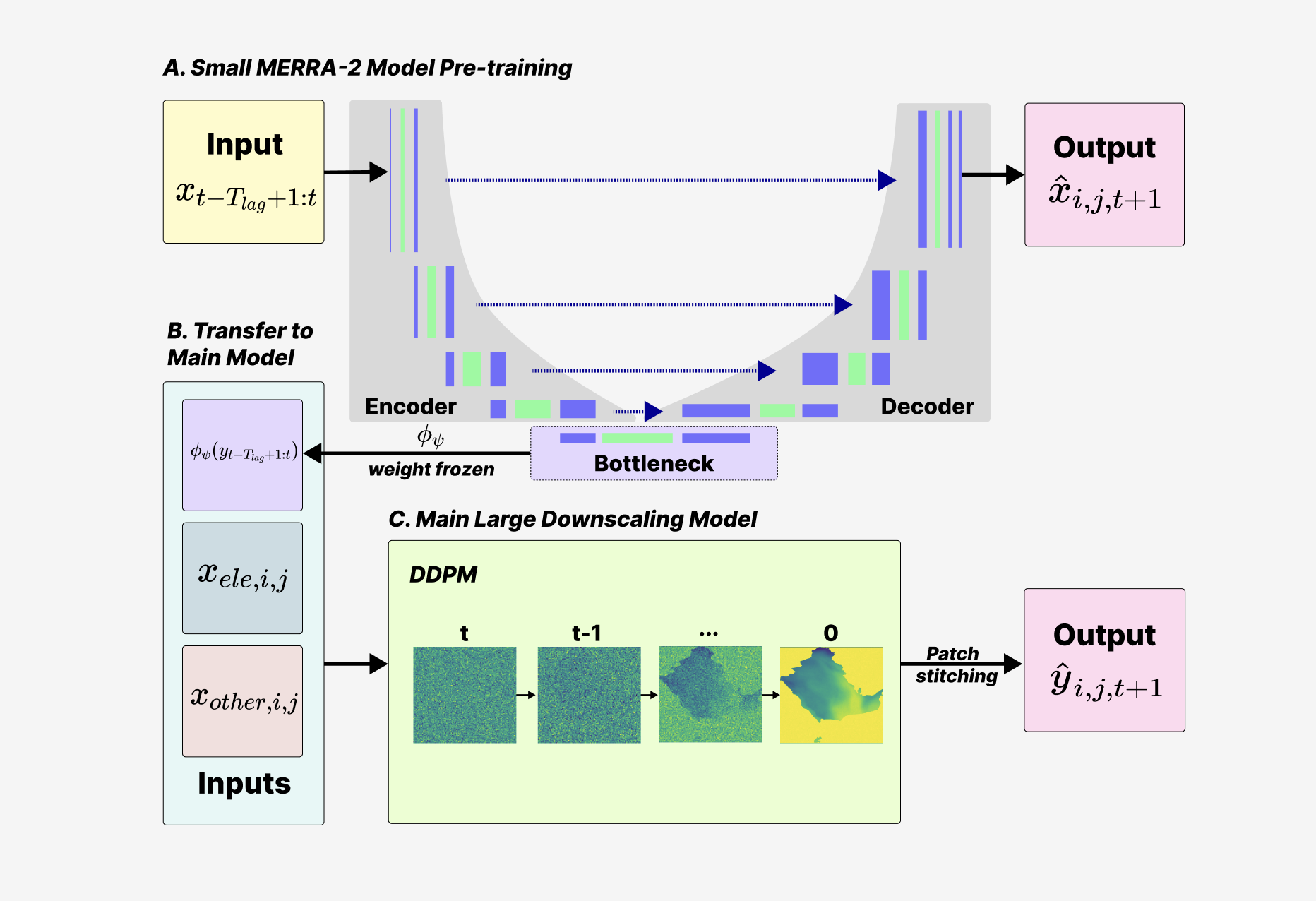}
    \caption{Overview of the downscaling framework:  
    (A) A small U-Net model \(f^{\mathrm{UNet}}_{\Omega}\) is pretrained on MERRA-2 to learn long-range spatiotemporal structure; (B) The encoder $\phi_{\psi}$ is frozen from model (A), extracted, and transferred to generate inputs for G5NR sequences; (C) The large downscaling denoising diffusion probabilistic model \(f^{\mathrm{DDPM}}_{\Theta}\) predicts high-resolution G5NR dust-extinction image $\hat y_{i,j,t+1}$ for day $t{+}1$.}
    \label{fig:ddpm-pipeline}
\end{figure}

The overarching goal of our modeling framework is to downscale a long time series of a target variable (AOD dust extinction) from coarse resolution images (MERRA-2, $\sim$ 50 km) based on their spatial relationship with fine resolution images (G5NR, $\sim$ 7 km) where there is a limited time series of overlapping data to learn from. This task is accomplished in two parts: %For each season and region, our downscaling model is composed of two parts. 
First, a simpler, small U-Net was trained only on the coarse resolution images to learn the spatiotemporal structure of the target variable across a long time series of images.
%MERRA-2 data to learn the long-time range temporal-spatial pattern of dust extinction. 
From this model, weights are encoded representing the extracted  knowledge of MERRA-2, and then transferred to the larger downscaling model as an encoded feature. The large model includes these transferred encoded features, along with MERRA-2 for day \(t{+}1\), elevation,  that images' corresponding latitudes, longitudes, a seasonal index, the number of days since the first day in the dataset, and the normalized year information, to predict G5NR dust-extinction for day \(t{+}1\). Figure \ref{fig:ddpm-pipeline} illustrates the high-level structure of our downscaling framework.

We compare this downscaling framework to two other approaches, both leveraging the small transfer learning model, but replacing the large downscaling model with a deterministic U-Net or a variational autoencoder (VAE).

% define notations and dimensioanlity of the datasets. Also, define the input and output of our large model.

\subsubsection{Small Transfer Learning Model}
% Description of the Small Model on MERRA-2

Let \(x_{i,j,t}\) be dust-extinction in MERRA-2, whose spatial resolution is unified with that of G5NR, at latitude and longitude grid index \((i,j)\) and day \(t\); let \(y_{i,j,t}\) be the G5NR value at the same latitude and longitude grid index \((i,j)\) and day \(t\).  Also, the temporal sequences of MERRA-2 with our pre-specified task lag \(T_{lag}\) are denoted as \(\mathbf{x}_{t-\Lag+1:t}\) and the corresponding G5NR sequences are denoted as \(\mathbf{y}_{t-\Lag+1:t}\). Because G5NR provides only two years of data, training a downscaling model solely on its limited temporal sequences risks overfitting and poor generalization beyond its training period. To address this limitation, we first train $f_{\Omega}$ to autoregressively predict our target variable one day ahead $t{+}1$ from the preceding $\Lag$ days:
\begin{equation}
x_{i,j,t+1} = f_{\Omega}\big(\mathbf{x}_{t-\Lag+1:t}\big).
\label{eq:small-model}
\end{equation}
This small model learns long-range spatiotemporal dependencies in MERRA-2 dust extinction that are difficult to capture from G5NR alone. After it is trained, its encoder layers $\phi_{\psi}$ are weight-frozen and transferred to the main downscaling model $f_{\Theta}$ as a fixed feature extractor, while decoder layers are discarded. 
%The encoded features produced by \(\phi_{\psi}\) are bilinearly interpolated to match the spatial resolution of the large downscaling model’s inputs. These interpolated features are concatenated with the MERRA-2 day-\(t{+}1\) inputs, elevation, and other geographic variables before being passed into the large model \(f_{\Theta}\). 
The transferred encoder provides a rich latent representation that captures spatiotemporal patterns in dust extinction and guides high-resolution prediction in $f_{\Theta}$.

Each small model shares an autoencoder-like structure, consisting of an encoder that maps the temporal input sequence into a latent feature space and a decoder that reconstructs or predicts the next time step. The structure above applies to all small models with different parameterized \(f_{\Omega}\). For large downscaling models \(f^{\mathrm{UNet}}_{\Theta}\) and \(f^{\mathrm{DDPM}}_{\Theta}\), their associated small pre-trained model is a deterministic U-Net \(f^{\mathrm{UNet}}_{\Omega}\). The small pre-trained model for \(f^{\mathrm{VAE}}_{\Theta}\) is a deterministic Autoencoder model with LSTM \(f^{\mathrm{LSTM}}_{\Omega}\).

% introduction to definition of Transfer learning, negative transfer and catastrophic forgetting, and our transfer learning model.

This pretraining and encoder-transfer design follows the general paradigm of transfer learning, where knowledge obtained from a data-rich source domain is reused to enhance learning in a data-scarce target domain. To formalize this framework, we define the source and target domains as follows.
Let the source domain in the transfer learning model be 
\(D_S = (\mathcal{X}_s, P_s)\), and the target domain be 
\(D_T = (\mathcal{X}_t, P_t)\). 
Here, let \(\mathcal{X}_s\) be the feature space of the source domain, 
and \(P_s\) be the probability distribution over the source feature space. 
Moreover, for a specific source domain, let 
\(T_s = (\mathcal{Y}_s, f_s(y \mid x))\) 
be the task of the source domain. 
\(\mathcal{Y}_s\) is the label space of the source domain, 
which contains all possible true outputs that the model 
\(f_s(y \mid x)\) tries to predict. 
Transfer learning is defined as the process of enhancing the learning 
of a predictive function \(f_s(y \mid x)\) in a \emph{target domain} \(D_T\) 
by leveraging knowledge obtained from a \emph{source domain} \(D_S\) 
and its associated task \(T_S\) \citep{5288526}. 
This setting applies when either the data distributions differ 
(\(D_S \neq D_T\)) or the learning tasks are not identical 
(\(T_S \neq T_T\)). Under the context of transfer learning used in this research, MERRA-2 is the source domain, and G5NR is our target domain. 

% explain the potential drawbacks of transfer learning and how they impact our model: negative transfer and catastrophic forgetting
However, transfer learning might potentially suffer from negative transfer and catastrophic forgetting under various contexts \citep{5288526, NEURIPS2019_c6bff625}. The negative transfer refers to the phenomenon that the performance of the model's learning in the target domain is compromised by the transferred source and target data \citep{5288526}. In other words, the performance of the model trained on both source and target data is inferior to that of the model trained on target data alone, which could happen if the domain divergence between source and target domains is huge \citep{5288526}. Furthermore, catastrophic forgetting refers to the tendency of the main model to overfit on the target domain while abruptly forgetting the knowledge learned on the source domain, which could be caused by an aggressive fine-tuning of the pre-trained model with target domain information \citep{NEURIPS2019_c6bff625}. Under the context of this research, negative transfer would occur if the domain similarity between the MERRA-2 and G5NR domains is so low that knowledge transferred from MERRA-2 impaired the model’s ability to predict G5NR. Catastrophic forgetting would happen if the spatiotemporal knowledge learned from the long-term MERRA-2 were not effectively retained in the main model during transfer, and our main model would fail to borrow the advantages of MERRA-2’s extended temporal coverage. Mitigating both negative transfer and catastrophic forgetting is therefore essential for the design of our model to ensure robust and meaningful downscaling in this research.

\subsubsection{Main Downscaling Model}
We transfer the frozen encoded weights to the larger DDPM downscaling model (Figure~\ref{fig:ddpm-pipeline}) and the comparison models. Additional data preparation steps include unifying the coarse and fine resolution grids. For each geographical grid cell \((i,j)\) and day \(t\) in the particular region and season, 
let \(x_{i,j,t}\) be dust-extinction in MERRA-2, whose spatial resolution is unified with that of G5NR, at latitude and longitude grid index \((i,j)\) and day \(t\); let \(y_{i,j,t}\) be the G5NR value at the same latitude and longitude grid index \((i,j)\) and day \(t\). Also, the temporal sequences of MERRA-2 with our pre-specified task lag \(T_{lag}\) are denoted as \(\mathbf{x}_{t-\Lag+1:t}\) and the corresponding G5NR sequences are denoted as \(\mathbf{y}_{t-\Lag+1:t}\).
Let \(x_{\mathrm{ele},\,i,j}\) denotes elevation at latitude and longitude value of \((i,j)\) and \(x_{\mathrm{other},\,i,j}\) denotes other geographic variables used in the large model.  The transferred weights are applied to the G5NR sequences \(\mathbf{y}_{t-\Lag+1:t}\) to produce encoded features for G5NR. Bilinear interpolation is applied to the encoded feature to align the output dimension of \(\phi_{\psi}\) with the larger model's input data dimension. The large model \(f_{\Theta}\) then uses all inputs and transferred encoded features to predict \(y_{i,j,t+1}\). Altogether, our downscaling model can be expressed as the following:

\begin{equation}
\label{eq:main-reg}
y_{i,j,t+1}
\;=\;
f_{\Theta}\!\Big(
x_{i,j,t+1},\;
x_{\mathrm{ele},\,i,j},\;
x_{\mathrm{other},\,i,j},\;
\underbrace{\phi_{\psi}\!\big(\mathbf{y}_{t-\Lag+1:t}\big)}_{\text{transfer feature}}\
\Big)
\end{equation} %

Finally, a sigmoid activation function is applied at the large model's output to ensure the prediction range is the same as our normalized data.

The formula above applies to all large downscaling models with different \(f_{\Theta}\), and each parameterized main model corresponds to a different \(f_{\Theta}\): our proposed DDPM \(f^{\mathrm{DDPM}}_{\Theta}\) that directly predicts \(y_{i,j,t+1}\) under a squared-cosine noise schedule \citep{Nichol2021ImprovedDD}, and comparison models including a supervised U-Net \(f^{\mathrm{UNet}}_{\Theta}\) trained with a regression loss, and a Variational Autoencoder (VAE) \(f^{\mathrm{VAE}}_{\Theta}\) trained on a hybrid loss that combines KL Divergence and regression loss. It is also important to clarify that while the U-Net model deterministically predict \(y_{i, j, t+1}\), the VAE and the DDPM are probabilistic models that define a distribution and report a prediction for \(y_{i, j, t+1}\) with the mean of its predicted distribution. 
%The full model pipeline of \(f^{\mathrm{DDPM}}_{\Theta}\) is shown in Figure~\ref{fig:ddpm-pipeline}.

% explain and justify how our model avoid catastrophic forgetting

Our model pipeline addresses potential problems of the Transfer Learning in the following ways. First, it has been shown and proven that when a pre-trained model is transferred into a target task, maintaining its parameters close to their pre-trained values can effectively prevent catastrophic forgetting \cite{NEURIPS2019_c6bff625, Li2018ExplicitIB}. Following this principle, the encoder of our MERRA-2 pre-trained model is reused as a fixed feature extractor in the large downscaling model, and its weights are frozen and not updated during training on the target domain. Compared to other transfer learning methods, such as fine-tuning, weight freezing significantly reduces the computational burden of updating weights, while also avoiding catastrophic forgetting by maintaining pre-trained weights. Thus, weight freezing preserves the pre-trained weights and the rich spatiotemporal knowledge learned from the long-term MERRA-2 data, which prevents the main model from overfitting or forgetting the transferred knowledge.

% explain and justify how our model avoid negative transfer
Second, the risk of negative transfer in our model is mitigated by the inherent similarity between the source (MERRA-2) and target (G5NR) domains. Both datasets contain the same variable dust extinction over identical geographical regions using NASA’s atmospheric simulation models \citep{daSilva2014}. Thus, our model assumes that the domain divergence between the source and target domains is minimal, and transferring knowledge from MERRA-2 to G5NR will not compromise our target task. A quantitative verification of this assumption is provided later in Section \ref{sec:domain-similarity}, where we compute the Wasserstein distance between the normalized MERRA-2 and G5NR dust-extinction domains across all regions and seasons \citep{9938381}. Note that, however, this domain similarity metric is not a part of any training loss in our downscaling models.

\subsection{Training}
\label{sec:training}

Because we have two distinct regions in downscaling target countries and natural seasonal splitting, we train one model per (\emph{region}, \emph{season}) to ensure regional heterogeneity and seasonal dynamics in our trained models. For each region \(r\) and season \(s\), we create particular training and testing datasets:
\[
\mathcal{D}_{r,s}=\Big\{(\mathbf{x}_{t-\Lag+1:t},\,\mathbf{y}_{t-\Lag+1:t},\,x_{t+1},\,y_{t+1},\,x_{\mathrm{ele}},\,x_{\mathrm{other}})\;\Big|\; \text{day }t{+}1\text{ falls in season }s \text{ of region } r\Big\},
\]
Specifically, we extract data from all days within the same seasonal range across \emph{all years} available in the datasets and their corresponding elevations and additional geographical variables. Note that our seasonal split is year-agnostic. For example, data from December 2005 is grouped together with data from December 2006 in the same seasonal dataset. Nevertheless, cross-annual variability is still retained in our model through including the variable normalized year in \(x_{\mathrm{other}}\). The same seasonal and regional split is applied to the training of both the small model on MERRA-2 and the large downscaling model. All models are optimized with \emph{Adam} with base learning rate $1{\times}10^{-6}$ and 30 epochs. Moreover, the loss function of our model is 
\begin{equation}
\label{eq:loss}
\mathcal{L}(\Theta)
\;=\;
\underbrace{\mathcal{L}_{\text{data}}(\Theta)}_{\text{task loss}}
\;+\;
\lambda_{\mathrm{wd}}\,
\underbrace{\sum_{p\in\mathcal{P}_{\text{train}}} \|p\|_2^2}_{\text{weight decay}}
\end{equation}
where $\mathcal{P}_{\text{train}}$ are the set of all \emph{trainable} parameters of the model. For the loss of the large downscaling model, the parameters from the transferred encoder $\phi_\psi$ is frozen and not included in the regularization term of weight decay. $\mathcal{L}_{\text{data}}$ is the model-specific task. The pixel-wise MAE loss is used for all models except for \(f^{\mathrm{VAE}}_{\Theta}\). For VAE, KL Divergence with a scaled image reconstruction loss is used. As shown in Equation~\eqref{eq:loss}, all models are regularized by weight decay with hyperparameter $\lambda_{\mathrm{wd}}$ to prevent overfitting. 

Furthermore, additional regularizations are adopted. We further reserve $20\%$ of the training days as a train validation data used for monitoring model fitting and imposing regularization. To strictly prevent data leakage of assigning shuffled patches of pixels from the same day into both training and validation datasets, we perform the data splitting \emph{at the day level} and use group-aware shuffling in the data loaders. All patches derived from the same day are treated as a single group so they appear either in the training stream \emph{or} in the validation/callback stream within an epoch, never both. Reduce-on-plateau scheduling is used to monitor the model performance on train validation data by reducing the learning rate by a pre-set proportion if the model continuously shows no improvement on train validation data after a patience window $\kappa$. Moreover, early stopping is used to halt training when the training validation metric fails to improve for a pre-set amount of patience window, even if training iterations have not met our pre-specified epoch. We also save the best-performing model through checkpoints on the train validation data.  

\subsection{Inference and Patch Stitching.}

At inference we predict the day \(t{+}1\) dust–extinction image at the G5NR resolution by sliding \(16{\times}16\) patches with stride \(s{=}2\) across the image. A patch is identified by its top-left pixel \((y_0,x_0)\) and covers rows \(y\in[y_0,y_0{+}H)\) and columns \(x\in[x_0,x_0{+}W)\), with \(H{=}W{=}16\). Let \(\tilde{Y}_i\in\mathbb{R}^{H\times W}\) denote the network’s predicted dust–extinction patch for day \(t{+}1\) at patch index \(i\).

To reduce edge effects we keep only a core from each patch and drop a halo of \(h\) pixels on each side for interior patches. For patches that touch the full image boundary, the halo on the touching side is set to zero so the core remains inside the domain:
\[
h_i^{\text{top}}=\begin{cases}0,& y_0=0\\ h,& \text{otherwise}\end{cases},\quad
h_i^{\text{bottom}}=\begin{cases}0,& y_0{+}H=H_{\text{img}}\\ h,& \text{otherwise}\end{cases},
\]
\[
h_i^{\text{left}}=\begin{cases}0,& x_0=0\\ h,& \text{otherwise}\end{cases},\quad
h_i^{\text{right}}=\begin{cases}0,& x_0{+}W=W_{\text{img}}\\ h,& \text{otherwise}\end{cases}.
\]
The retained core index set is
\begin{equation}
\label{eq:core}
\mathcal{Q}_i=\big\{(y,x):\;
y_0{+}h_i^{\text{top}}\le y<y_0{+}H{-}h_i^{\text{bottom}},\;\;
x_0{+}h_i^{\text{left}}\le x<x_0{+}W{-}h_i^{\text{right}}\big\}.
\end{equation}

Because the halo \(h\) removes border pixels, the effective core is \((H-2h)\times(W-2h)\). To compensate, we use \(s{=}2\) for dense overlap. This averages each pixel over multiple overlapping cores and reduces variance and stitching artifacts, at the cost of evaluating more patches.

Before stitching we apply a separable two-dimensional Hann taper over each core \citep{10.1371/journal.pone.0229839}. Define the one-dimensional Hann window
\begin{equation}
\label{eq:hann1d}
w(n;N)=\tfrac{1}{2}\!\left(1-\cos\!\frac{2\pi n}{N-1}\right),\qquad n=0,1,\dots,N{-}1,
\end{equation}
where \(N\) is the window length. For patch \(i\), the core has
\[
N_i^{\text{row}} = H - h_i^{\text{top}} - h_i^{\text{bottom}},\qquad
N_i^{\text{col}} = W - h_i^{\text{left}} - h_i^{\text{right}}.
\]
For a core pixel \((y,x)\in\mathcal{Q}_i\), define its row and column indices within the core as
\[
n_y = y - (y_0{+}h_i^{\text{top}}),\qquad
n_x = x - (x_0{+}h_i^{\text{left}}).
\]
The 2D weight is
\begin{equation}
\label{eq:hann2d}
W_i(y,x)=w\!\big(n_y; N_i^{\text{row}}\big)\;w\!\big(n_x; N_i^{\text{col}}\big).
\end{equation}

We maintain two full-image accumulators. The weighted-sum image
\begin{equation}
\label{eq:sum}
S(y,x)=\sum_{i}\mathbf{1}\!\big\{(y,x)\in\mathcal{Q}_i\big\}\,W_i(y,x)\,\tilde{Y}_i(y,x)
\end{equation}
adds each core prediction with its weight, and the weight image
\begin{equation}
\label{eq:weight}
Z(y,x)=\sum_{i}\mathbf{1}\!\big\{(y,x)\in\mathcal{Q}_i\big\}\,W_i(y,x)
\end{equation}
adds the corresponding weights. The stitched prediction is
\begin{equation}
\label{eq:normalize}
\hat{Y}(y,x)=\frac{S(y,x)}{\max\{Z(y,x),\varepsilon\}},
\end{equation}
where \(\varepsilon\) is a small constant for numerical safety. Here \(\tilde{Y}_i(y,x)\) is the model’s patch-level prediction for day \(t{+}1\) at pixel \((y,x)\) inside patch \(i\), \(S(y,x)\) is the accumulated weighted sum from all overlapping patch cores, \(Z(y,x)\) is the accumulated total weight, and \(\hat{Y}(y,x)\) is the final stitched image at the G5NR resolution.

We evaluate patches in mini-batches to improve inference efficiency. Let \(\{\mathcal{B}_1,\dots,\mathcal{B}_M\}\) be any partition of the patch index set. For batch \(b\), after computing \(\{\tilde{Y}_i\}_{i\in\mathcal{B}_b}\) we update, for every pixel \((y,x)\) in the full image,
\[
\begin{aligned}
S(y,x) &\leftarrow S(y,x) \;+\; 
\sum_{i\in\mathcal{B}_b} \mathbf{1}\!\big\{(y,x)\in\mathcal{Q}_i\big\}\, W_i(y,x)\,\tilde{Y}_i(y,x),\\[2pt]
Z(y,x) &\leftarrow Z(y,x) \;+\; 
\sum_{i\in\mathcal{B}_b} \mathbf{1}\!\big\{(y,x)\in\mathcal{Q}_i\big\}\, W_i(y,x).
\end{aligned}
\]
We then proceed to the next batch until all patches are processed. Because these updates are additive, the stitched image \(\hat{Y}\) in \eqref{eq:normalize} is identical regardless of batch size or patch order; batching only affects computational efficiency. Combined with halo cropping and the Hann taper, this accumulation suppresses visible seams and blockiness.

For each Season$\times$Area split and day \(t{+}1\), our large model predicts the dust–extinction AOD image at the G5NR resolution. The inputs at the G5NR resolution are: (i) the MERRA\mbox{-}2 sequence \(\mathbf{x}_{t-\Lag+1:t}\); (ii) the MERRA\mbox{-}2 image for day \(t{+}1\) as the driver; (iii) transferred encoded features from the fixed encoder \(\phi_{\psi}\) applied to a G5NR sequence \(\mathbf{y}_{t-\Lag+1:t}\); and (iv) the geographic and calendar variables used elsewhere in this work, namely elevation \(x_{\mathrm{ele},\,i,j}\), latitude, longitude, the season index, the number of days since the first day in the datasets, and the normalized year. All inputs use \(\log_{10}\) dust–extinction AOD and are standardized with the mean and standard deviation computed from the training days of the same Season$\times$Area split; the same parameters are reused for validation, test, and out-of-data periods. Predictions are mapped back to data units by reversing the standardization and the \(\log_{10}\) transform.

The same prediction pipeline is used in the 2005–2007 overlap period and after 2007, with only the source of the encoder sequence differing. In the overlap period, we form the encoder input with the true G5NR sequence \(\mathbf{y}_{t-\Lag+1:t}\). After 2007, when true G5NR is unavailable, we use an autoregressive update at G5NR resolution: initialize the encoder input with the last \(\Lag\) \emph{true} G5NR images from the overlap period; for each subsequent day \(u\), predict \(\hat{y}_{i,j,u}\), append \(\hat{y}_{\cdot,\cdot,u}\), and drop the oldest image to form \(\hat{\mathbf{y}}_{u-\Lag+1:u}\) for \(\phi_{\psi}\). The driver and all other inputs remain MERRA\mbox{-}2 at the G5NR resolution together with the same geographic and calendar variables.

\subsection{Evaluation}\label{sec:Evaluaion}

\subsubsection{Domain-Similarity Diagnostics}
\label{sec:domain-similarity}

% compute wasserstein distance
To assess the degree of divergence between the source (MERRA-2) and target (G5NR) domains, we compute the 1-Wasserstein distance between their normalized dust-extinction fields. This diagnostic is used to confirm the validity of the transfer-learning assumption introduced in Section \ref{sec:modeling}. The Wasserstein distance between the source and target domains is computed as the following. Specifically, for a fixed day \(t\) and region,
let \(\Omega_t\subseteq\{1,\dots,H\}\times\{1,\dots,W\}\) denote the set of pixel indices \((i,j)\)
that are inside the region and valid in both fields on day \(t\).
We apply a common log–min–max normalization to both maps using shared bounds over \(\Omega_t\).
Define
\begin{align}
b_x(i,j) &= \log_{10}\!\big(\max(x_{i,j,t},\varepsilon)\big), &
b_y(i,j) &= \log_{10}\!\big(\max(y_{i,j,t},\varepsilon)\big),
\end{align}
with \(\varepsilon>0\) to avoid \(\log(0)\) and \((i,j)\in\Omega_t\).
Compute joint bounds
\[
m_t \;=\; \min_{(i,j)\in\Omega_t}\{\,b_x(i,j),\,b_y(i,j)\,\}, \qquad
M_t \;=\; \max_{(i,j)\in\Omega_t}\{\,b_x(i,j),\,b_y(i,j)\,\}.
\]
Normalize each map:
\begin{equation}
\tilde{x}_{i,j,t}=\operatorname{clip}_{[0,1]}\!\left(\frac{b_x(i,j)-m_t}{M_t-m_t}\right), \qquad
\tilde{y}_{i,j,t}=\operatorname{clip}_{[0,1]}\!\left(\frac{b_y(i,j)-m_t}{M_t-m_t}\right),
\label{eq:wd-norm}
\end{equation}
where \(\operatorname{clip}_{[0,1]}(u)=\min\{1,\max\{0,u\}\}\) truncates values to \([0,1]\).

\noindent
Let \(F_{\tilde{x}_t}\) and \(F_{\tilde{y}_t}\) be the empirical CDFs of
\(\{\tilde{x}_{i,j,t} : (i,j)\in\Omega_t\}\) and \(\{\tilde{y}_{i,j,t} : (i,j)\in\Omega_t\}\).
The 1-Wasserstein distance is
\begin{equation}
W_1(\tilde{x}_t,\tilde{y}_t)
\;=\;
\int_{0}^{1} \bigl| F_{\tilde{x}_t}(u) - F_{\tilde{y}_t}(u) \bigr| \,\mathrm{d}u.
\label{eq:wd}
\end{equation}
For an all-day summary, we use a histogram approximation on \([0,1]\) with \(B\) equal-width bins.
If \(\hat F_{\tilde{x}_t}\) and \(\hat F_{\tilde{y}_t}\) are the corresponding empirical cumulative histograms, we use Riemann sum approximation with equal bin width of \(\frac{1}{B}\) to approximate the integral: 
\begin{equation}
\widehat{W}_1(\tilde{x}_t,\tilde{y}_t)
\;=\;
\frac{1}{B}\sum_{k=1}^{B}
\Bigl|\hat F_{\tilde{x}_t}(k/B)-\hat F_{\tilde{y}_t}(k/B)\Bigr|.
\label{eq:wd-hist}
\end{equation}

\subsubsection{In-Data Downscaling Performance}\label{sec:in-data evaluation}
We evaluate the in-data downscaling performance for each seasonal–regional model separately using the corresponding test samples. All evaluation metrics are computed in a pixel-wise manner across spatial grids to assess local accuracy and spatial consistency between model predictions and the true data.

We validate the in-data downscaling prediction from both the small model and the large model using the test samples. For the small transfer model under the DDPM framework, the predictions are transformed by $log_{10}$ and normalized during training. In the evaluation process, we collected the denormalized predictions $\hat{x}_{i,j,t}$ and the true MERRA-2 data $x_{i,j,t}$. We then calculate the mean absolute error (MAE) to reflect the mean deviation between predicted and observed values, and the Coefficient of Determination ($R^2$) to measure the proportion of variance in the observed data that is explained by the model on the $log_{10}$ transformed data 
$$
\begin{aligned}
\text{MAE} &= \frac{1}{N} \sum_{i,j} \left| x_{i,j,t} - \hat{x}_{i,j,t} \right|
&
\text{R}^2 &= 1 - \frac{\sum_{i,j} \left(x_{i,j,t} - \hat{x}_{i,j,t} \right)^2}{\sum_{i,j} \left(x_{i,j,t} - \overline{x_{t}} \right)^2}
\end{aligned}
$$
The smaller the value of MAE is, the better the model is. The theoretical range for $R^2$ is from negative infinity to 1, where 1 represents that the model perfectly explains the variability in the data, 0 means the model is no better than just using the mean of the observations, and negative values indicate performance even worse.\\
For the large UNet3D model, the normalized and $log_{10}$-transformed predictions correspond to G5NR dust extinction. We load both the denormalized predictions $\hat{y}_{i,j,t}$ and the true G5NR data $y_{i,j,t}$. Then compute the Root Mean Squared Error (RMSE) and Mean Absolute Error (MAE) to quantify overall magnitude and bias of residuals, where a smaller value for both MAE and RMSE indicates better model performance, RMSE gives higher weight to large errors compared to MAE, and same to MAE.
$$
\begin{aligned}
\text{MAE} &= \frac{1}{N} \sum_{i,j}| y_{i,j,t} - \hat{y}_{i,j,t}|
&
\text{RMSE} &= \sqrt{\frac{1}{N} \sum_{i,j} \left(y_{i,j,t} -\hat{y}_{i,j,t} \right)^2}
\end{aligned}
$$
Coefficient of Determination ($R^2$) is also calculated, as well as the Nash–Sutcliffe Efficiency (NSE), which evaluates predictive skill relative to the observed mean:
$$\text{NSE} = 1 - \frac{\sum_{i,j} \left(y_{i,j,t} - \hat{y}_{i,j,t} \right)^2}{\sum_{i,j} \left(y_{i,j,t} - \overline{y_t} \right)^2}$$
Like $R^2$, the range of NSE is from negative infinity to 1, with higher values denoting better performance. However, NSE measures predictive skill relative to the observed mean.
We also adopt the Kling–Gupta Efficiency (KGE) in the evaluation of the large model, which decomposes the model performance into correlation ($r$), bias ratio ($\beta$), and variability ratio ($\gamma$), with a higher KGE value indicating better model:
$$\text{KGE} = 1 - \sqrt{(r-1)^2 + (\beta-1)^2 + (\gamma-1)^2}$$ where $r$ is the Pearson correlation coefficient $$r = \frac{\sum_{i,j} (y_{i,j,t} - \overline{y_t}) (\hat{y}_{i,j,t} - \overline{\hat{y}_t})}{\sqrt{\sum_{i,j} (y_{i,j,t} - \overline{y_t})^2} \sqrt{\sum_{i,j} (\hat{y}_{i,j,t}- \overline{\hat{y_t}})^2}}$$ $\beta$ is the bias ratio $\beta = \frac{\overline{\hat{y}_t}}{\overline{y_t}}$ , and $\gamma$ is the variability ratio $\gamma = \frac{\sigma(\hat{y}_{i,j,t})}{\sigma(y_{i,j,t})}$ where $\sigma(y_{i,j,t})$ is the standard deviation of true values $y_{i,j,t}$, and $\sigma(\hat{y}_{i,j,t})$ is the standard deviation of predicted value $\hat{y}_{i,j,t}$. By integrating accuracy, bias, and variability within a single framework, KGE provides a more balanced measure of model behavior, and it is more robust when a model reproduces the spatial-temporal pattern of the data but differs in magnitude or variance.\\
For each test day, metrics are first computed for every pixel in the spatial domain to assess individual accuracy. The pixel-level scores are then averaged over all spatial grid cells to obtain a daily mean metric. Finally, the daily metrics are averaged over all test days to summarize the overall seasonal–regional performance. Temporal alignment between predictions and targets was verified, and data integrity checks were implemented to ensure consistent evaluation across all regions and seasons.\\
\subsubsection{Out-of-Data Downscaling Performance}\label{sec:out-of-data evaluation}
To examine how well the diffusion-based downscaling method preserves physical structure outside the training period, we evaluated both spatial and temporal dependencies of the predicted dust extinction variable (log-transformed) in comparison to the MERRA-2 data and the high-resolution truth G5NR.\\
Because the true G5NR data are unavailable for the forecast period, quantitative validation cannot be performed directly. Instead, we rely on graphical and statistical diagnostics such as semivariogram and temporal autocorrelation analysis to assess whether the predicted spatial–temporal variability remains physically plausible.
These diagnostics are well-suited for out-of-data evaluation, as they do not require point-to-point correspondence with ground truth but instead measure the statistical consistency of spatial structures and temporal dynamics with the historical reference patterns observed in G5NR and MERRA-2. Their validity was first confirmed in the in-data downscaling experiment, where the semivariogram and temporal metrics showed a clear and physically consistent relationship between true G5NR and the true MERRA-2 data. This establishes that these graphical measures can reliably capture the cross-scale physical linkage, justifying their use for evaluating out-of-data predictions.\\
For spatial structure analysis, semivariograms were computed daily and seasonally aggregated. For each Season-Area combination, the semivariance for each dataset $z \in \{x,y,\hat{y}\}$ \[
\gamma_z(h) = \tfrac{1}{2} E\left[\big(z(s,t) - z(s+h,t)\big)^{2}\right],
\] where $z(s)$ denotes the log-transformed dust extinction at spatial location $s$, and $h$ is the spatial distance between any two predicted grid cells (pixels). By computing the rate at which the average squared difference between pixel values increases with $h$, we quantify the decay of spatial correlation. We estimate semivariance from 30,000 random pixel pairs binned by great-circle distance up to 600 km. A spherical model was fitted to derive nugget, sill, and range parameters. During the overlap period, comparing $\gamma_{\hat{y}}(h)$ with $\gamma_{y}(h)$ confirms that the model successfully reproduces the correct spatial variance spectrum of true G5NR. In OOD years, we compare $\gamma_{\hat{y}}(h)$ with the physically consistent reference $\gamma_{\hat{x}}(h)$, agreement between these curves indicates that the prediction maintains realistic spatial correlation structure in the absence of ground truth.\\ \\
To evaluate temporal dependencies in both the training and out-of-data periods, we employed two approaches: the lag-based image-wise statistics including RMSE and $R^2$ and the autocorrelation function (ACF) and partial autocorrelation function (PACF).\\
For each Season–Area combination, we reduce data into regional mean:
$$
\begin{aligned}
\bar{x}_t&=\frac{1}{HW}\sum_{i,j}x_{i,j,t}
&
\bar{y}_t&=\frac{1}{HW}\sum_{i,j}y_{i,j,t}
&
\bar{\hat{y}}_t&=\frac{1}{HW}\sum_{i,j}\hat{y}_{i,j,t}
\end{aligned}
$$
For each $\overline{z}_t$, we compute both the autocorrelation function (ACF)
\[
\rho_{\overline{z}}(k) =
\frac{\operatorname{Cov}\big(\overline{z}_t, \overline{z}_{t-k}\big)}
{\text{Var}({\overline{z}_t})}.
\]
and the partial autocorrelation function (PACF) up to 30-day lags for the model predictions. The ACF measures overall linear dependence between observations $k$ days apart, whereas the PACF removes the indirect effects of shorter lags, capturing only the direct correlation between $\overline{z}_t$ and $\overline{z}_{t-k}$ after removing the effect of all intermediate lags 1, ..., k-1.
Together, they describe the persistence and effective memory length of the predicted data. During the overlap period, if $\rho_{\overline{\hat{y}}}(k)$ closely matches $\rho_{\overline{y}}(k)$, then it validates that the model captures the correct temporal memory characteristics of G5NR. In OOD years, we compare $\rho_{\overline{y}}(k)$ to the reference $\rho_{\overline{x}}(k)$, the predicted series should remain within the physical range established by historical MERRA-2 and G5NR.\\ \\
In parallel, we quantify the temporal coherence among the datasets. We computed image-wise RMSE and $R^2$ between day $t$ and day $t-\ell$ for lags $\ell=1{\ldots}10$ days.
$$
\begin{aligned}
    \text{RMSE}_z(\ell)&=\sqrt{\frac{1}{HW}\sum_{i,j}(z_{i,j,t}-z_{i,j,t-\ell})^2}
    &    
    \text{R}^2_z(\ell)&=1-\frac{\sum_{i,j}(z_{i,j,t}-z_{i,j,t-\ell})^2}{\sum_{i,j}(z_{i,j,t}-\overline{z})^2}
\end{aligned}
$$
Smaller RMSE and larger $R^2$ at short lags indicate stronger temporal persistence, whereas rapid growth in RMSE and decline in  $R^2$ with increasing lag reflect faster decorrelation of spatial structures. In OOD years, predictions are expected to produce lag-RMSE and lag-$R^2$ curves lying within the interval of historical MERRA-2 and G5NR behaviors.\\
Together, the semivariogram, ACF/PACF, and lag-based RMSE/$R^2$ curves provide a comprehensive assessment of out-of-data performance. They verify whether the predicted high-resolution fields $\hat{y}_{i,j,t}$ preserve the spatial variance, temporal persistence, and decorrelation behavior associated with true G5NR and MERRA-2. These diagnostics enable credible validation even in periods where no high-resolution ground truth is available.
% ============================ 3. Result =========================
\section{Result}\label{sec:Result}

\subsection{Exploratory Data Analysis}\label{sec:Exploratory Data Analysis (EDA)}

\begin{figure}[H]
    \centering
    \includegraphics[width=0.9\textwidth]{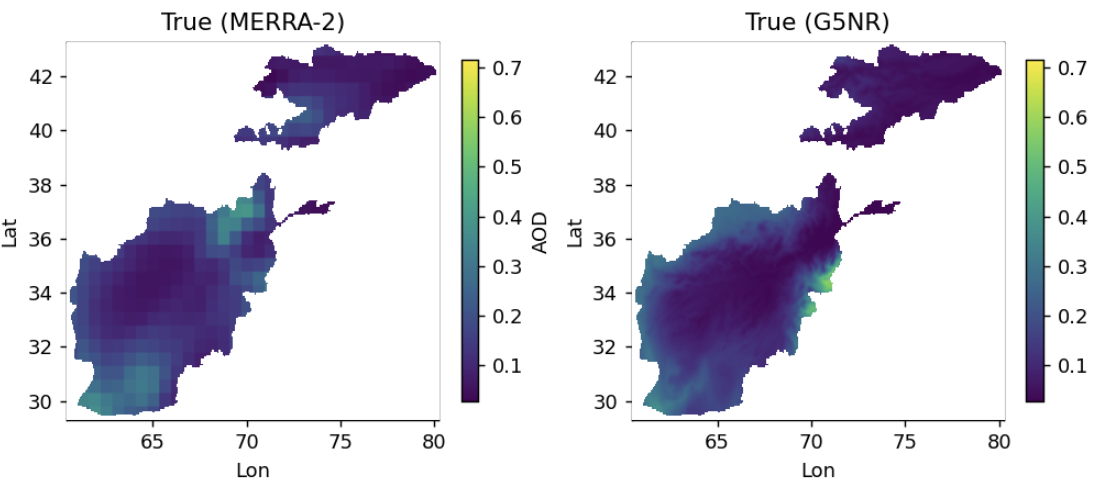}\\[1em]
    \includegraphics[width=0.9\textwidth]{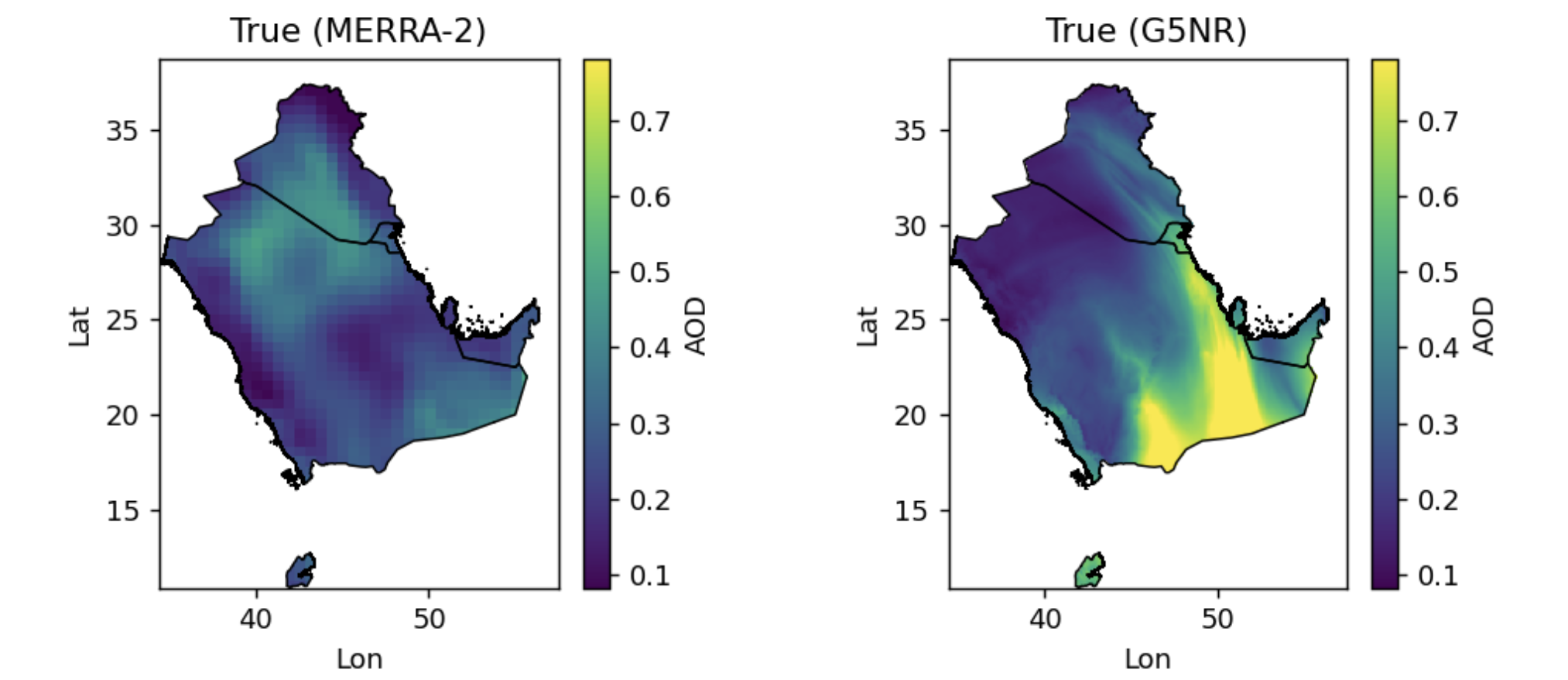}
    \caption{Top: image of true G5NR (left) and MERRA-2 (right) chosen from a single day in Season 3 and Area 0 (Afghanistan and Kyrgyzstan) with their corresponding latitude and longitude in y and x axes. Bottom: image of single-day true G5NR (left) and MERRA-2 (right) chosen from season 3 and Area 1 (Gulf countries and Djibouti)}
    \label{fig:true_predicted_images}
\end{figure}

Figure~\ref{fig:true_predicted_images} visualizes true images from both MERRA-2 and G5NR for a single day selected from the two modeling regions during the summer season (June-August). Across both regions, the MERRA-2 images appear visually blocky, which reflects their coarse spatial resolution and limited ability to capture fine-scale landscape features. In contrast, the corresponding G5NR images exhibit substantially finer spatial resolution and reveal more nuanced spatial variability and dynamic dust extinction patterns across the terrain. Moreover, the G5NR images often preserve finer details within broader structural patterns that are also present in MERRA-2.  For example, both images indicate higher dust extinction values in the southeastern part of Saudi Arabia and the southwestern part of Afghanistan. However, spatial pattern discrepancies also exist between MERRA-2 and G5NR. In Djibouti, for instance, the MERRA-2 image shows relatively moderate dust extinction values, whereas G5NR depicts high intensities. We also observe high dust extinction intensities at southern Iraq and Northern Saudi Arabia from MERRA-2, whereas G5NR shows low intensities at those regions.

\begin{figure}[H]
    \centering
    % --- Top row: R^2 plots ---
    \begin{subfigure}[t]{0.48\textwidth}
        \centering
        \includegraphics[width=\textwidth]{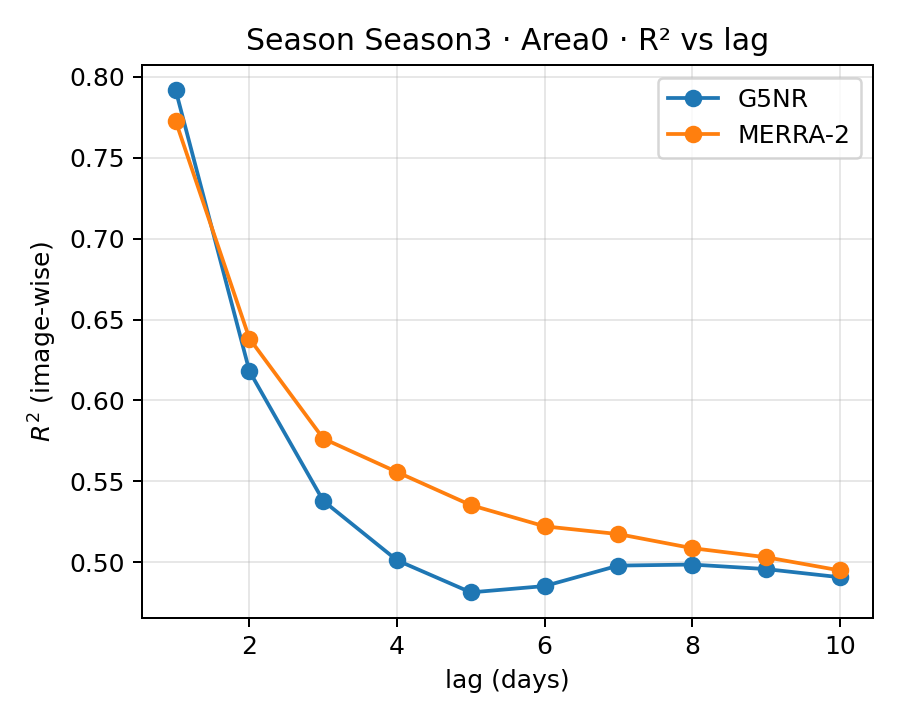}
        \caption{Imagewise $R^2$ from Season~3, Area~0}
    \end{subfigure}
    \hfill
    \begin{subfigure}[t]{0.48\textwidth}
        \centering
        \includegraphics[width=\textwidth]{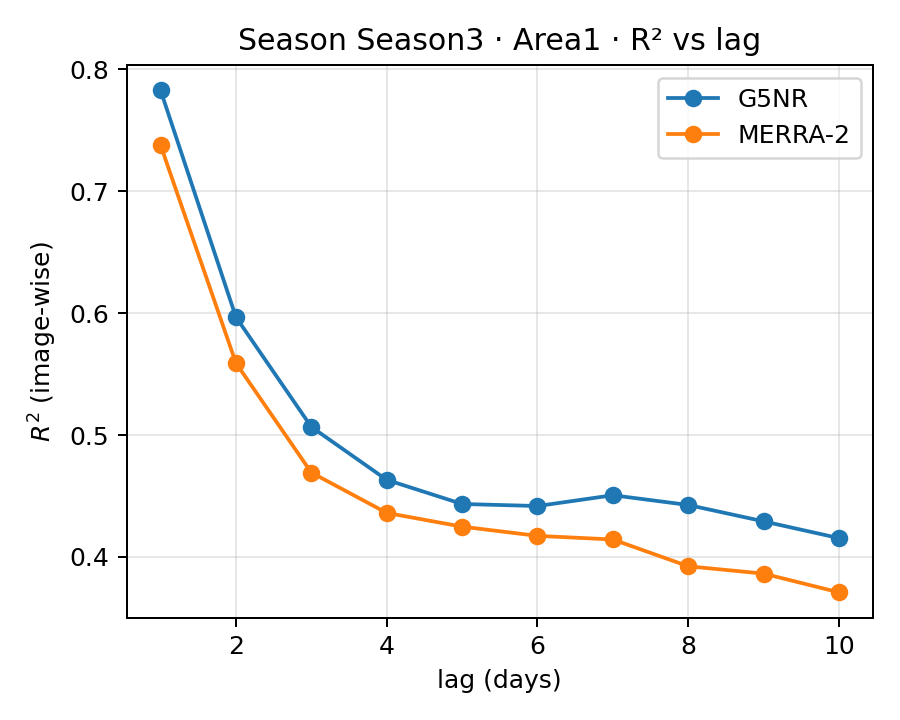}
        \caption{Imagewise $R^2$ from Season~3, Area~1}
    \end{subfigure}

    \vspace{1em}

    % --- Bottom row: RMSE plots ---
    \begin{subfigure}[t]{0.48\textwidth}
        \centering
        \includegraphics[width=\textwidth]{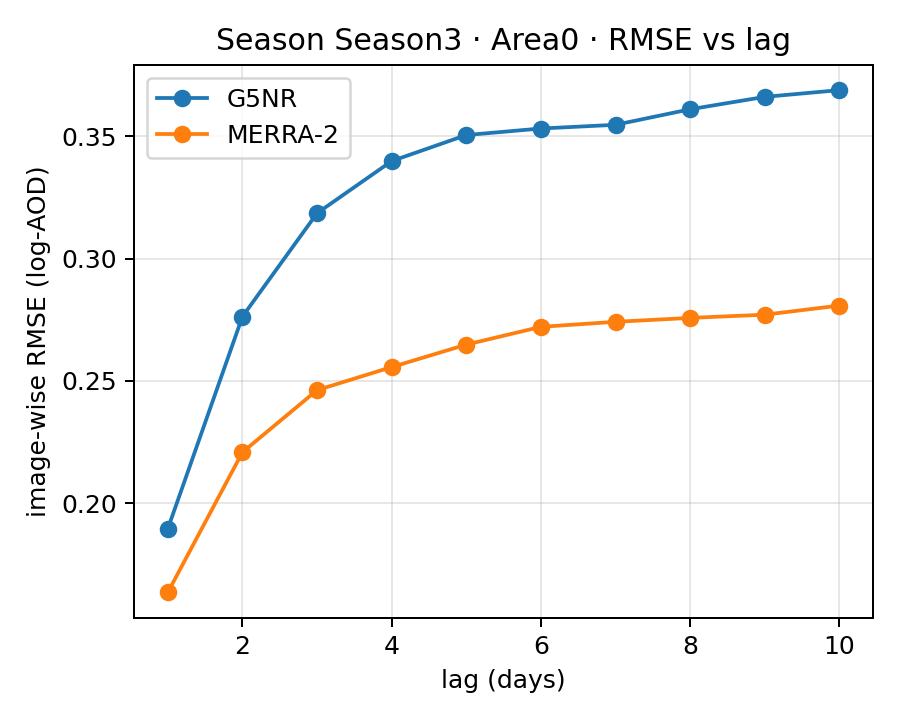}
        \caption{Imagewise RMSE from Season~3, Area~0}
    \end{subfigure}
    \hfill
    \begin{subfigure}[t]{0.48\textwidth}
        \centering
        \includegraphics[width=\textwidth]{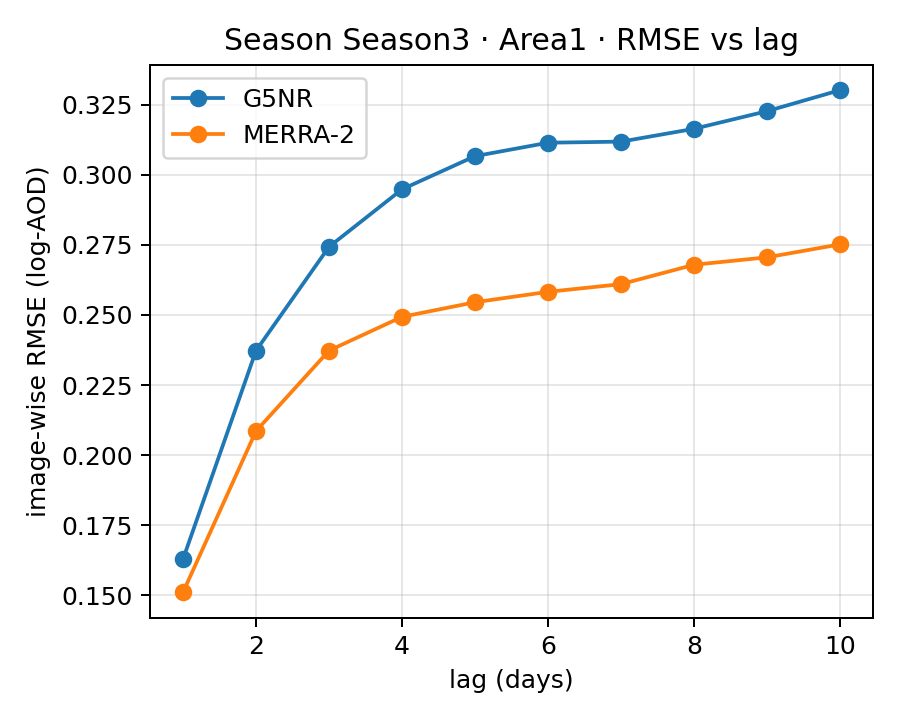}
        \caption{Imagewise RMSE from Season~3, Area~1}
    \end{subfigure}

    \caption{Pixel-wise temporal-lag diagnostics for Season~3 in both regions. 
    Each panel shows within-dataset temporal persistence of true G5NR and MERRA-2 dust-extinction fields. 
    Top row: pixel-wise $R^2$ between daily fields separated by 1–10~day lags; 
    bottom row: corresponding RMSE values. 
    Higher $R^2$ indicates stronger temporal persistence, while increasing RMSE reflects faster temporal variability among images across time lags.}
    \label{fig:pixel_lags_images}
\end{figure}

Further graphical diagnostics for G5NR and MERRA-2 dust extinction is presented in Figure~\ref{fig:pixel_lags_images}. 
For each region in Season 3, these plots show the average pixel-wise $R^2$ (top panels) and RMSE (bottom panels) between all images separated by a temporal lag of 1–10 days within the same dataset. The $x$-axis indicates the lag in days, where a lag of 1 corresponds to consecutive days ($t$ and $t\!-\!1$). Higher $R^2$ values indicate stronger temporal persistence of spatial patterns, whereas increasing RMSE reflects faster temporal changes in dust-extinction magnitudes. 
Across both regions, $R^2$ decreases sharply as lag increases, and RMSE rises correspondingly, which indicates that their image spatial structures decorrelate over roughly a one-week window.
G5NR consistently exhibits higher RMSE than MERRA-2 at the same lag, suggesting that G5NR captures finer-scale and more transient atmospheric dynamics, whereas MERRA-2 evolves more smoothly over time due to its coarser resolution.
Despite these magnitude differences, the two datasets display highly similar overall graphical trends, implying that their underlying temporal associations are highly similar and comparable. We observed similar patterns for other seasons as well. Therefore, the shared patterns provide a valuable diagnostic tool for out-of-data downscaling: when G5NR ground truth is unavailable but MERRA-2 remains available, the MERRA-2 pixel-wise lag metric curves can serve as a physically consistent reference for validating whether the predicted G5NR fields exhibit realistic temporal coherence and evolution over successive days.

The summary statistics for seasonal and regional dust extinction from G5NR and MERRA-2 are presented in the Table~\ref{tab:summary-statistics}. 
Although both datasets represent the same dust extinction variable, G5NR generally exhibits higher mean values than of MERRA-2. This difference is expected given the algorithmic distinctions between the two simulation systems \citep{daSilva2014}. Furthermore, the wider spread in quantiles and larger standard deviations in G5NR indicate greater spatial variability, which aligns with its finer spatial resolution and enhanced ability to capture detailed variation and dynamics across the landscape.

% Table of Summary Statistics

\FloatBarrier
\begin{table}[H]
\centering
\caption{Summary statistics for the dust extinction variable in G5NR and MERRA-2 across each season and area, computed over overlapping years in the original dataset units.}
\label{tab:summary-statistics}

\unifiedtabsetup
\resizebox{\linewidth}{!}{%
\begin{tabular}{l c c c c c c c c}
\toprule
& \multicolumn{2}{c}{\shead{Season 1}} & \multicolumn{2}{c}{\shead{Season 2}} & \multicolumn{2}{c}{\shead{Season 3}} & \multicolumn{2}{c}{\shead{Season 4}} \\
\cmidrule(lr){2-3}\cmidrule(lr){4-5}\cmidrule(lr){6-7}\cmidrule(lr){8-9}
\textbf{Statistic / Region} & \theadsm{A0} & \theadsm{A1} & \theadsm{A0} & \theadsm{A1} & \theadsm{A0} & \theadsm{A1} & \theadsm{A0} & \theadsm{A1} \\
\midrule
\multicolumn{9}{l}{\textbf{G5NR}}\\
$n_{\text{days}}$ & \num{270} & \num{270} & \num{291} & \num{291} & \num{292} & \num{292} & \num{272} & \num{272} \\
Mean              & \num{0.0811} & \num{0.1478} & \num{0.1890} & \num{0.2624} & \num{0.2502} & \num{0.3684} & \num{0.1623} & \num{0.2757} \\
Std               & \num{0.0966} & \num{0.1430} & \num{0.2049} & \num{0.2354} & \num{0.2579} & \num{0.2697} & \num{0.1913} & \num{0.2512} \\
5 Quantile        & \num{0.0046} & \num{0.0103} & \num{0.0108} & \num{0.0184} & \num{0.0322} & \num{0.0548} & \num{0.0157} & \num{0.0289} \\
50 Quantile       & \num{0.0503} & \num{0.1090} & \num{0.1277} & \num{0.1940} & \num{0.1599} & \num{0.3117} & \num{0.0931} & \num{0.2125} \\
95 Quantile       & \num{0.2595} & \num{0.4194} & \num{0.5664} & \num{0.7419} & \num{0.7683} & \num{0.8773} & \num{0.5868} & \num{0.7391} \\
\addlinespace[3pt]
\multicolumn{9}{l}{\textbf{MERRA-2}}\\
$n_{\text{days}}$ & \num{270} & \num{270} & \num{291} & \num{291} & \num{292} & \num{292} & \num{272} & \num{272} \\
Mean              & \num{0.0667} & \num{0.1157} & \num{0.1510} & \num{0.2184} & \num{0.1835} & \num{0.2958} & \num{0.1164} & \num{0.1889} \\
Std               & \num{0.0633} & \num{0.1038} & \num{0.1409} & \num{0.1730} & \num{0.1657} & \num{0.1883} & \num{0.1300} & \num{0.1459} \\
5 Quantile        & \num{0.0072} & \num{0.0183} & \num{0.0228} & \num{0.0353} & \num{0.0341} & \num{0.0686} & \num{0.0127} & \num{0.0333} \\
50 Quantile       & \num{0.0499} & \num{0.0867} & \num{0.1168} & \num{0.1683} & \num{0.1449} & \num{0.2586} & \num{0.0855} & \num{0.1521} \\
95 Quantile       & \num{0.1792} & \num{0.3183} & \num{0.3804} & \num{0.5692} & \num{0.4527} & \num{0.6441} & \num{0.3172} & \num{0.4657} \\
\bottomrule
\end{tabular}%
}

\vspace{0.4em}
\centering
\begin{minipage}{0.9\linewidth}
\scriptsize
\emph{Notes:} Statistics are computed over all pixels in each seasonal and regional subset using the original untransformed data units. 
A0 denotes Region~0 (Afghanistan and Kyrgyzstan), and A1 denotes Region~1 (Gulf countries and Djibouti). 
The higher standard deviation in G5NR compared with MERRA-2 reflects its finer spatial resolution and greater spatial variability.
\end{minipage}

\end{table}
\FloatBarrier

We summarize spatial and temporal dependence over the overlap period of True G5NR and True MERRA-2 (May 2005–June 2007). All arrays use $\log_{10}$ dust–extinction AOD. Computations follow the Season$\times$Area split. For spatial structure, we compute an empirical semivariogram at each dataset’s native resolution for each day and then average the daily curves within the split; a spherical fit is overlaid on the averaged curve \citep{gmd-15-2505-2022}. For temporal structure, we form the area-mean daily series within the split and compute ACF and PACF for lags 1–30 days \citep{31540}.

\begin{figure}[H]
  \centering
  \begin{subfigure}[t]{0.48\textwidth}
    \centering
    \includegraphics[width=\textwidth]{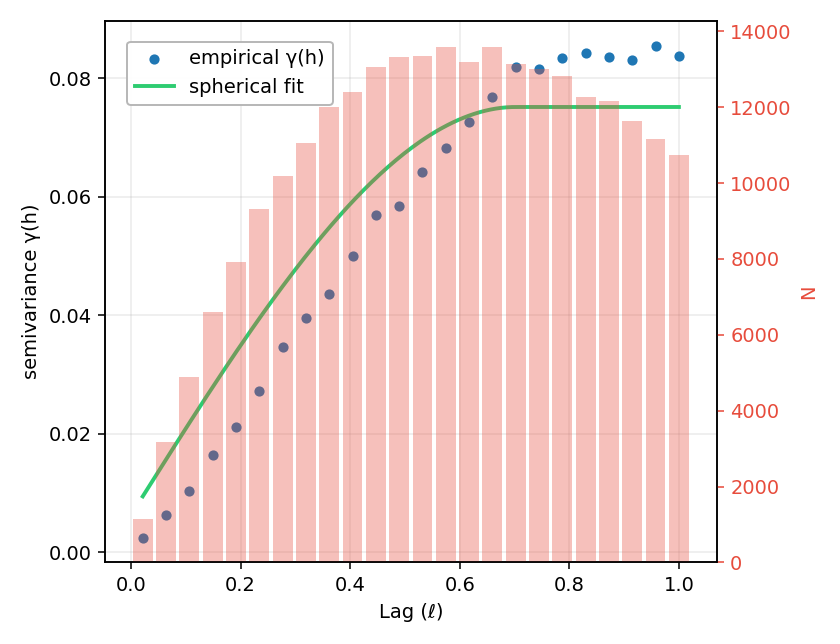}
    \caption{Season~3, A0. True G5NR semivariogram. Daily curves are averaged over all days in the split; bars show pair counts per distance bin; spherical fit overlaid. Higher sill and shorter range reflect stronger small-scale variability.}
  \end{subfigure}\hfill
  \begin{subfigure}[t]{0.48\textwidth}
    \centering
    \includegraphics[width=\textwidth]{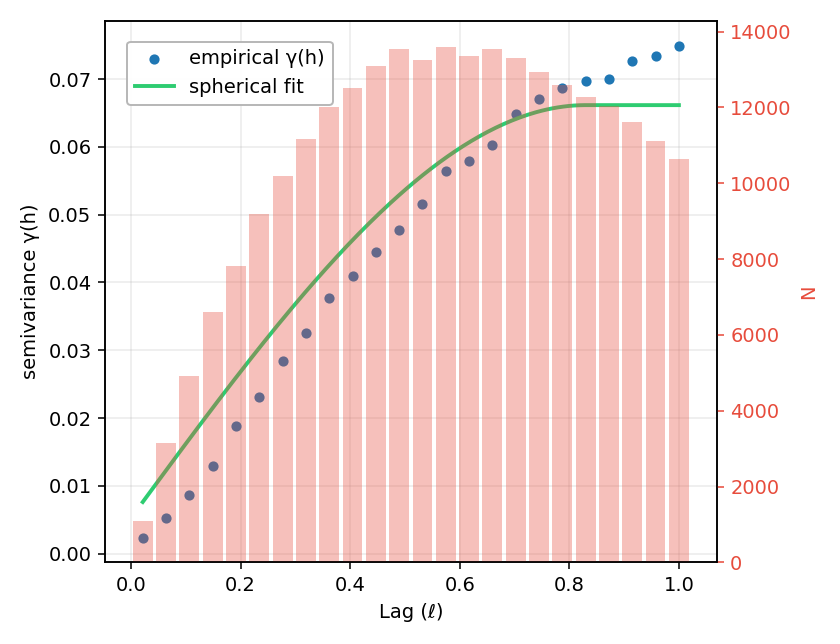}
    \caption{Season~3, A0. True MERRA-2 semivariogram. Averaged daily curve with pair-count bars and spherical fit. Lower sill and longer range indicate smoother arrays and longer spatial correlation.}
  \end{subfigure}
  \caption{Native-resolution spatial structure during the overlap period of True G5NR and True MERRA-2. Semivariograms are computed per day and averaged within the Season$\times$Area split; spherical fits summarize the averaged curves.}
  \label{fig:native_semivariograms}
\end{figure}

Figure~\ref{fig:native_semivariograms} shows Season~3, A0 semivariograms computed at each dataset’s native resolution for $\log_{10}$ dust–extinction AOD. The $x$-axis is the distance between two pixel centers within the Season$\times$Area split, grouped into bins from near 0 up to about 800\,km \citep{Karney2011AlgorithmsFG}. We compute a semivariogram for each day, average across days in the split, and summarize the mean curve with a spherical fit. At short pixel–pair distances the G5NR curve starts higher and rises more steeply than the MERRA-2 curve, indicating larger small-scale variance and sharper gradients at G5NR resolution. G5NR also reaches its sill at a shorter distance, so spatial correlation decays faster with increasing pixel–pair distance. MERRA-2, at a coarser resolution, rises more gently and has a longer range, consistent with stronger spatial smoothing. For distances of about 50–800\,km both curves level off at similar sill values, indicating comparable large-scale variance. In short, G5NR concentrates variance at short spatial scales, whereas MERRA-2 spreads variance over longer ranges.

\begin{figure}[H]
  \centering
  \includegraphics[width=0.92\textwidth]{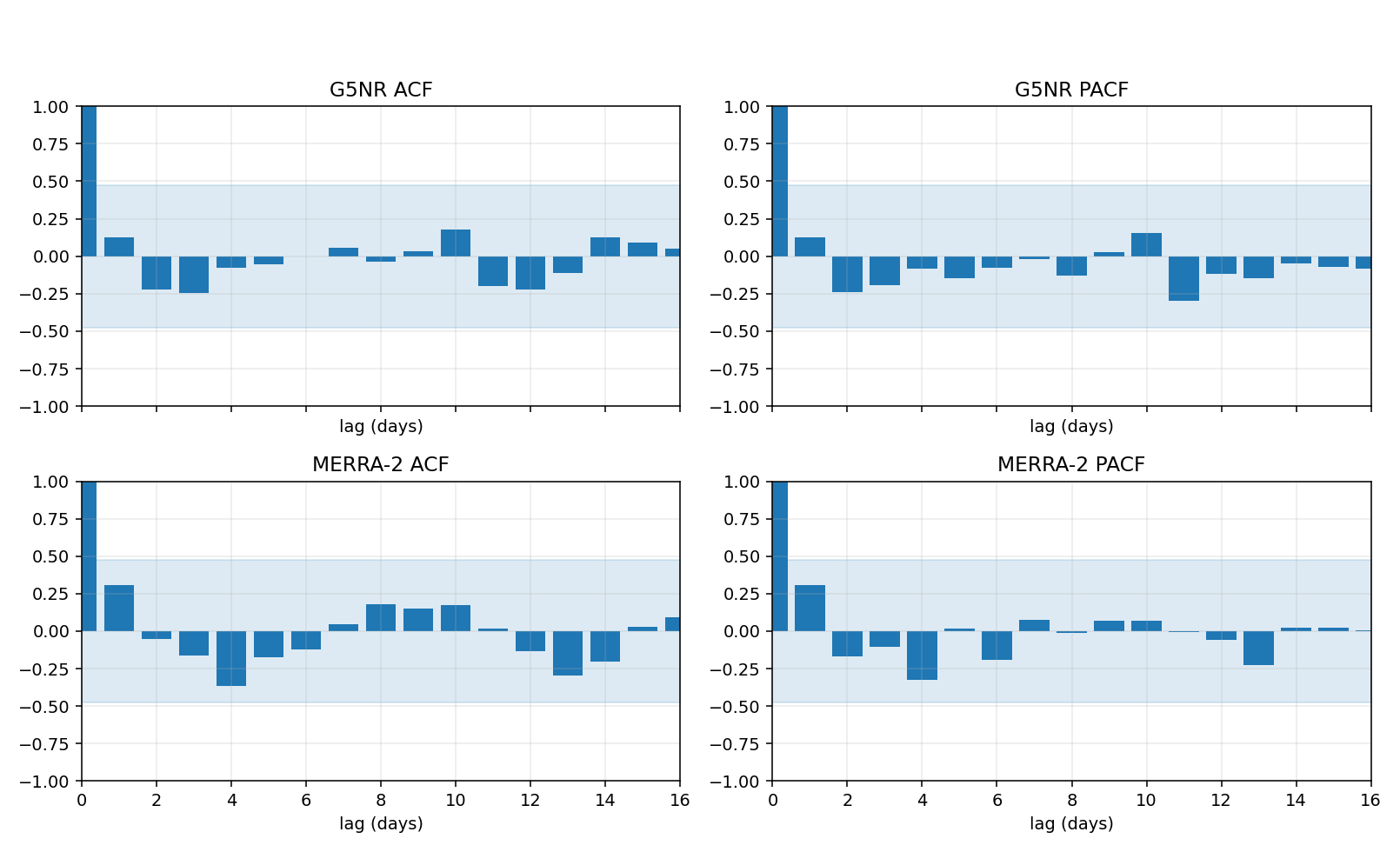}
  \caption{Native-resolution temporal structure during the overlap period for Season~3, A0. ACF and PACF of the area-mean AOD for lags 1–16 days, computed within the Season$\times$Area split. Faster ACF decay for G5NR indicates stronger short-lag variability; PACF mass concentrates at the first few lags for both datasets.}
  \label{fig:native_acf_pacf}
\end{figure}

Figure~\ref{fig:native_acf_pacf} shows, for Season~3 and A0, the ACF and PACF of a daily regional series constructed as follows: for each day, take dust extinction AOD at every valid pixel at each dataset’s native resolution, average these values over the region to obtain one number for that day, and compute ACF/PACF for lags 1–16. In both datasets the ACF stays positive for several short lags, indicating multi-day persistence. The G5NR ACF decays toward zero in fewer lags than MERRA-2, consistent with larger day-to-day variability at G5NR resolution. The PACF shows clear nonzero spikes at lags 1–2 (occasionally 3) and is near zero thereafter. Thus today’s regional average is influenced mainly by the last one to three days, with a shorter effective memory in G5NR than in MERRA-2.

\subsection{Domain-Similarity Diagnostics}
\label{sec:domain-similarity-results}

Table~\ref{tab:wd-similarity} presents the Wasserstein distance (WD) between the MERRA-2 and G5NR dust extinction domains. Since both MERRA-2 and G5NR domains are normalized, the range of the resulting WD will be within the range of \([0,1]\). The metric was calculated both per day over their overlapping period (763 days from May~2005 to June~2007) and globally over the entire temporal coverage of each dataset. We also compute WD for each region respectively, because our model addresses two regions the differently. Across both regions, the computed WD values are close to zero, demonstrating high domain similarity between the two datasets. For Region~0, the mean daily WD over the overlap period is \(0.0282\), with a 10th–90th percentile range of \([0.0114,\,0.0478]\), and a pooled WD of \(0.0080\) when comparing all available MERRA-2 and G5NR data. Similarly, for Area~1 (the Gulf countries and Djibouti), the mean daily WD is \(0.0237\) with percentile range \([0.0087,\,0.0371]\), and a pooled WD of \(0.0151\). 

These consistently low WD values indicate minimal domain shift between MERRA-2 and G5NR. Consequently, the assumption of domain similarity underlying our transfer learning design is empirically validated. This result justifies our use of a weight frozen parameter-based transfer without additional regularization terms designed in the training loss. Therefore, freezing the pre-trained encoder not only preserves transferable features learned from MERRA-2, but also provides a robust foundation for learning G5NR-specific spatiotemporal patterns without incurring negative transfer.

\FloatBarrier
\begin{table}[H]
\centering
\caption{Wasserstein distances (WD) between normalized MERRA-2 and G5NR dust-extinction fields. WD values close to zero indicate high domain similarity.}
\label{tab:wd-similarity}

\unifiedtabsetup
\begin{tabular}{l c c c}
\toprule
\textbf{Region} & \textbf{Daily WD Mean} & \textbf{Quantile Range} & \textbf{Pooled WD} \\
\midrule
Region 0 & 0.0282 & [0.0114,\,0.0478] & 0.0080 \\
Region 1 & 0.0237 & [0.0087,\,0.0371] & 0.0151 \\
\bottomrule
\end{tabular}

\vspace{0.4em}
\begin{minipage}{0.9\linewidth}
\scriptsize
\emph{Notes:} WD values are computed after applying log$_{10}$ transformation and joint min–max normalization over the \([0,1]\) range. Per-day WD metrics summarize 763 overlapping days between 2005–2007, while pooled WD compares the full temporal domains of both datasets. P10 and P90 mean the 10th and 90th quantiles for all computed per-day WD. Lower values of WD indicate higher domain similarity between MERRA-2 and G5NR and better justify the usage of transfer learning across two domains.
\end{minipage}
\end{table}
\FloatBarrier

\subsection{In-data Downscaling Results}\label{sec: in-data downscaling result under Result section}
Table~\ref{tab:result model-validation} presents the in-data downscaling performance of each model, evaluated on the held-out test days that were excluded from model training but for which true G5NR values are available. 
The VAE model exhibited signs of model collapse and unstable behavior. Although its RMSE values appear numerically small, sometimes smaller than those of the DDPM, it failed to capture the complex spatiotemporal dynamics present in the data. Its low $R^2$ values indicate that the model explains only a little variation in the true G5NR data and fail to capture the spatiotemporal association within G5NR. VAE's downscaled image is presented in Figure~\ref{fig:in_data_vae_prediction} in the Appendix. 

In contrast, both the U-Net and DDPM models achieved strong predictive performance, as reflected by their high $R^2$ values. The U-Net shows smaller values and standard deviations in RMSE compared with those of DDPM, which is expected because the U-Net is a deterministic model that produces a single point estimate, whereas the DDPM is a generative model that inherently exhibits higher variability across samples. However, the U-Net’s $R^2$ values fluctuate more noticeably than those of the DDPM, with performance deteriorating in Season~3 but recovering slightly by Season~4. 
The DDPM, by comparison, demonstrates greater performance stability and robustness across seasons and regions. Additional diagnostic metrics, such as the Nash–Sutcliffe Efficiency (NSE) and Kling–Gupta Efficiency (KGE), also confirm the strong in-data downscaling performance of the DDPM. More details of these computed metrics are provided in Table~\ref{tab:model-validation} in the Appendix.

% Table of Model Validation Results
\FloatBarrier
\begin{table}[H]
\centering
\caption{Validation performance of downscaling models evaluated on held-out G5NR test days across each season and area. Additional validation metrics are provided in Table~\ref{tab:model-validation} in the Appendix.}
\label{tab:result model-validation}

\unifiedtabsetup
{\normalsize
\setlength{\tabcolsep}{3.2pt}
\renewcommand{\arraystretch}{1.20}
\begin{adjustbox}{max width=\linewidth}
\begin{tabular}{l c c c c c c c c}
\toprule
& \multicolumn{2}{c}{\shead{Season 1}} & \multicolumn{2}{c}{\shead{Season 2}} & \multicolumn{2}{c}{\shead{Season 3}} & \multicolumn{2}{c}{\shead{Season 4}} \\
\cmidrule(lr){2-3}\cmidrule(lr){4-5}\cmidrule(lr){6-7}\cmidrule(lr){8-9}
\textbf{Model / Region} & \theadsm{A0} & \theadsm{A1} & \theadsm{A0} & \theadsm{A1} & \theadsm{A0} & \theadsm{A1} & \theadsm{A0} & \theadsm{A1} \\
\midrule

\multicolumn{9}{l}{\textbf{Large DDPM}}\\
$R^2$  & \cell{0.891}{0.068} & \cell{0.944}{0.032} & \cell{0.934}{0.034} & \cell{0.774}{0.082} & \cell{0.953}{0.020} & \cell{0.986}{0.006} & \cell{0.854}{0.046} & \cell{0.896}{0.059} \\
RMSE   & \cell{0.189}{0.053} & \cell{0.160}{0.031} & \cell{0.137}{0.058} & \cell{0.231}{0.049} & \cell{0.089}{0.020} & \cell{0.056}{0.014} & \cell{0.181}{0.037} & \cell{0.135}{0.028} \\
\addlinespace[2pt]
\multicolumn{9}{l}{\textbf{Large U-Net}}\\
$R^2$  & \cell{0.938}{0.017} & \cell{0.914}{0.040} & \cell{0.883}{0.063} & \cell{0.942}{0.026} & \cell{0.794}{0.106} & \cell{0.697}{0.114} & \cell{0.645}{0.143} & \cell{0.822}{0.088} \\
RMSE   & \cell{0.016}{0.003} & \cell{0.015}{0.001} & \cell{0.023}{0.008} & \cell{0.011}{0.002} & \cell{0.033}{0.010} & \cell{0.039}{0.008} & \cell{0.039}{0.018} & \cell{0.026}{0.008} \\
\addlinespace[2pt]
\multicolumn{9}{l}{\textbf{Large VAE}}\\
$R^2$  & \cell{0.065}{0.045} & \cell{0.437}{0.102} & \cell{0.098}{0.055} & \cell{0.360}{0.259} & \cell{0.010}{0.009} & \cell{0.019}{0.021} & \cell{0.063}{0.068} & \cell{0.100}{0.117} \\
RMSE   & \cell{0.102}{0.019} & \cell{0.095}{0.008} & \cell{0.149}{0.023} & \cell{0.090}{0.033} & \cell{0.326}{0.067} & \cell{0.181}{0.040} & \cell{0.133}{0.078} & \cell{0.143}{0.044} \\
\bottomrule
\end{tabular}
\end{adjustbox}
}

\vspace{0.4em}
\centering
\begin{minipage}{0.9\linewidth}
\scriptsize
\emph{Notes:} Model performance is evaluated using held-out G5NR data from the test days selected for model validation. For each row, the value without the bracket is the mean of the corresponding metric computed for all test days, and the value within the bracket is the standard deviation of all computed metrics for every test days. RMSE is computed in the same physical units as the original datasets.
\end{minipage}

\end{table}
\FloatBarrier

\begin{figure}[H]
    \centering
    \includegraphics[width=0.85\textwidth]{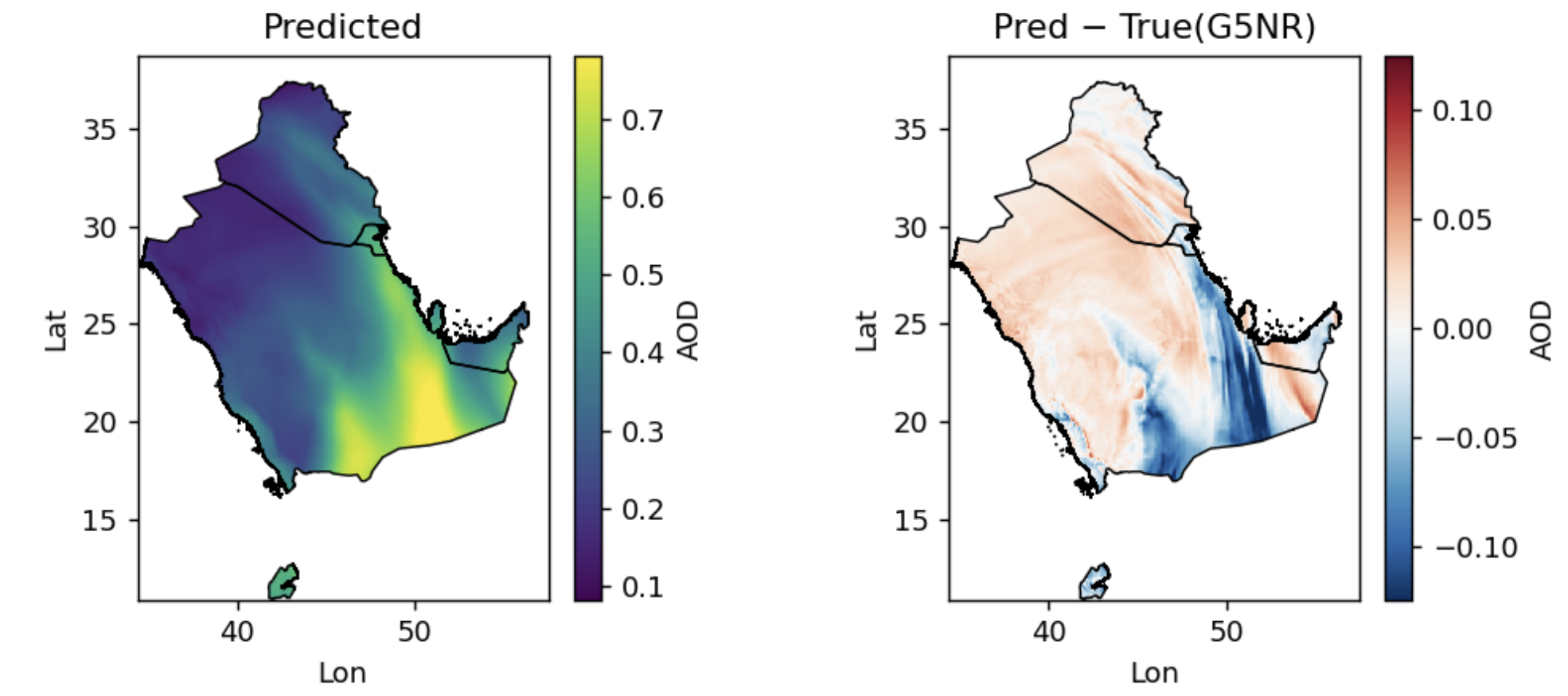}
    \caption{Visualization of in-data downscaling prediction (left) and pixel-wise difference between predicted and true G5NR dust extinction (right) for the same day from Season 3 shown in Figure~\ref{fig:true_predicted_images}. 
    The left panel shows the predicted G5NR produced by our model, while the right panel presents the difference between the predicted and true pixel values. For visualizations of other regions and seasons, please refer to Figure~\ref{fig:app_in_data_A0} in the Appendix.}
    \label{fig:in_data_prediction}
\end{figure}

The visualization of our prediction in Figure~\ref{fig:in_data_prediction} further confirms the excellent in-data downscaling performance summarized in Table~\ref{tab:model-validation}. The predicted image of AOD is highly consistent with the true G5NR observation visualized at the bottom-right of Figure~\ref{fig:true_predicted_images}, which indicates that the model successfully captures high-resolution spatial variability within the target region. We observed similar consistencies between true and predicted G5NR images at other regions and seasons. More details are presented in the Figure~\ref{fig:app_in_data_A0} in the Appendix.
Nevertheless, the pixel-wise difference plot reveals a slightly conservative behavior of the model: prediction errors tend to be larger in regions with extreme dust-extinction intensities compared with moderate-intensity areas. This suggests that while the model performs robustly overall, it slightly smooths sharp gradients associated with intense dust events.

\section{Out-of-data Downscaling Results}
\label{sec:ood-results}

We evaluate the large DDPM outside its training window by forecasting Season~3 of 2008 in A1. The model is initialized with the last \(\Lag\) days from the training period and then advances one day at a time in an autoregressive manner, driven only by the MERRA\mbox{-}2 day-\(t{+}1\) array. All analyses in this section use dust-extinction AOD in original scales, and the same Season\(\times\)Area split as elsewhere.

\begin{figure}[H]
  \centering
  \begin{subfigure}[t]{0.48\textwidth}
    \centering
    \includegraphics[width=\textwidth]{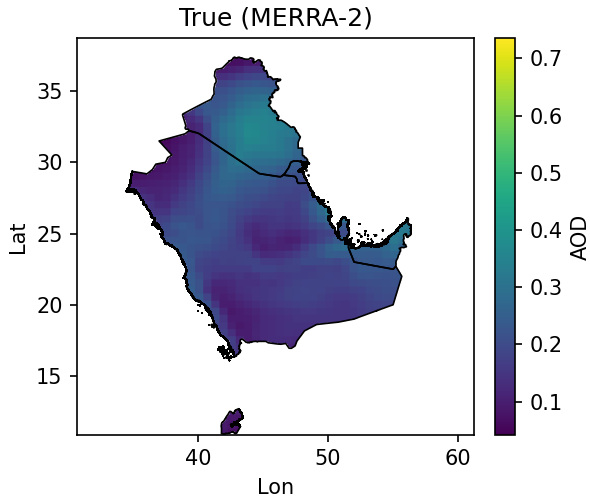}
    \caption{MERRA\mbox{-}2 at native resolution.}
  \end{subfigure}\hfill
  \begin{subfigure}[t]{0.48\textwidth}
    \centering
    \includegraphics[width=\textwidth]{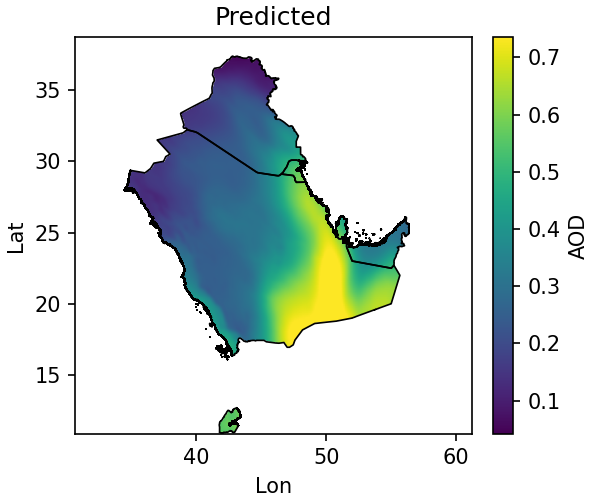}
    \caption{DDPM predicted at native G5NR resolution.}
  \end{subfigure}
  \caption{OOD comparison for 2008-07-11 (Season~3, A1). Left: driver from MERRA\mbox{-}2. Right: DDPM prediction. Both panels are shown at their native resolutions. The prediction follows the large-scale pattern in MERRA\mbox{-}2 and adds fine-scale plumes with sharper edges.}
  \label{fig:ood_merra_vs_pred}
\end{figure}

A representative OOD day (2008-07-11) illustrates the relation between the MERRA\mbox{-}2 driver and the DDPM output. Figure~\ref{fig:ood_merra_vs_pred} shows, left, the same day MERRA\mbox{-}2 and, right, the DDPM prediction at the native G5NR resolution. We do not have true G5NR for 2008, so the comparison is between driver and prediction only. The large-scale variations in MERRA-2 align with the broad structure in the prediction, while the prediction shows coherent fine-scale plumes and sharper edges. Both panels are shown at native resolution with no differencing, so the side-by-side view offers a simple check that the model keeps the large-scale daily patterns and adds realistic local detail.

\begin{figure}[H]
  \centering
  \begin{subfigure}[t]{0.48\textwidth}
    \centering
    \includegraphics[width=\textwidth]{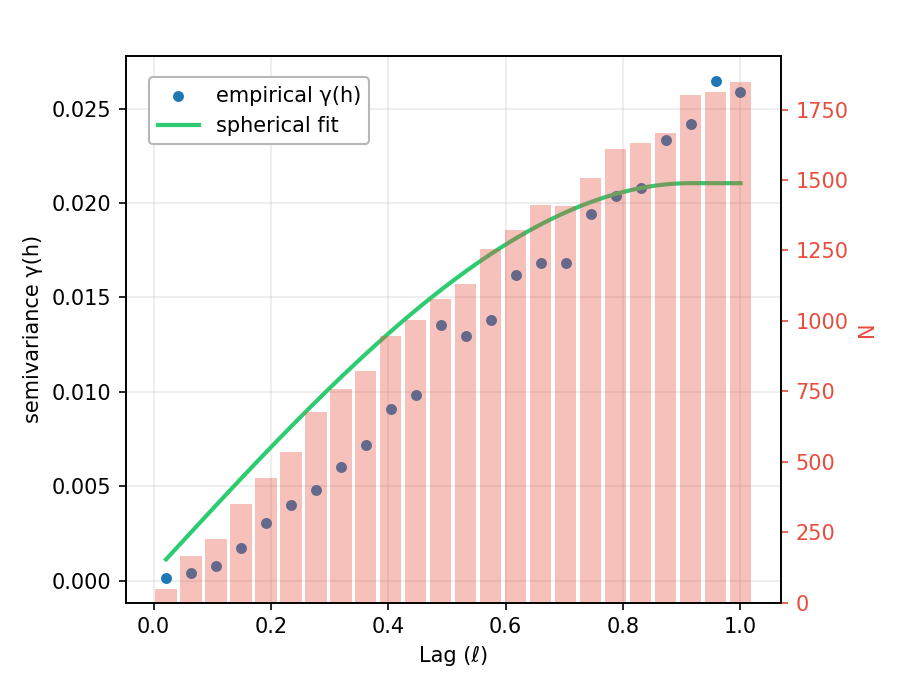}
    \caption{Semivariogram of predicted at native G5NR resolution, averaged over OOD days.}
  \end{subfigure}\hfill
  \begin{subfigure}[t]{0.48\textwidth}
    \centering
    \includegraphics[width=\textwidth]{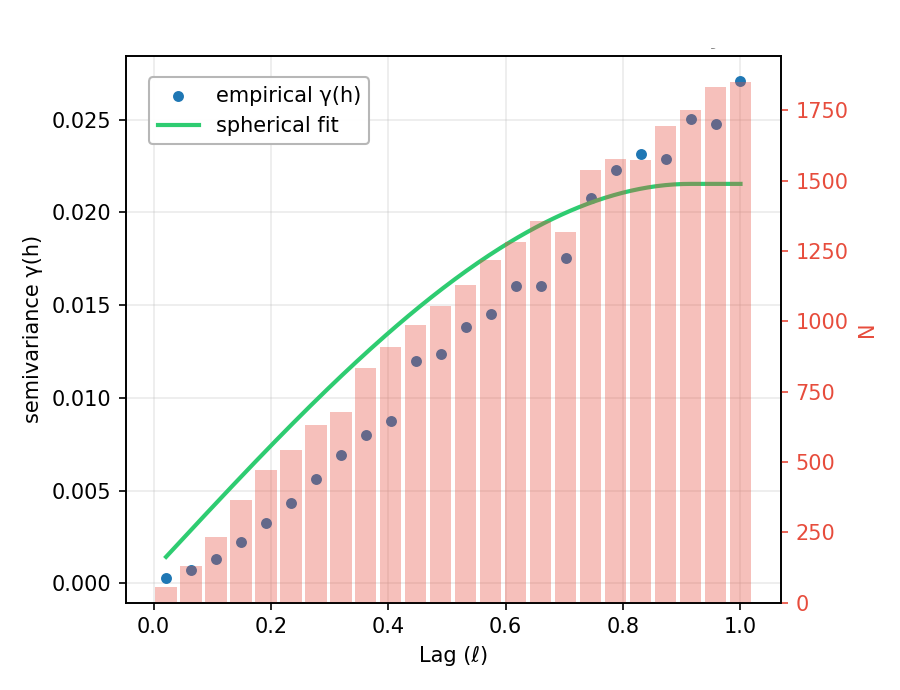}
    \caption{Semivariogram of MERRA\mbox{-}2 regridded to the G5NR resolution, averaged over the same days.}
  \end{subfigure}
  \caption{Spatial dependence in the OOD period (Season~3, A1, 2008). Predicted shows higher small-lag semivariance and a shorter correlation range than MERRA\mbox{-}2 regridded to G5NR resolution, while both approach similar sills at large lags.}
  \label{fig:ood_semivariogram}
\end{figure}

To quantify spatial behavior across Season~3, 2008, we compute daily isotropic empirical semivariograms for two series: (i) the prediction at native G5NR resolution and (ii) MERRA\mbox{-}2 regridded to the G5NR resolution for the same dates. Figure~\ref{fig:ood_semivariogram} aggregates these daily semivariograms. The prediction shows systematically higher semivariance at short lags together with a shorter fitted spherical range, indicating stronger small-scale variability and faster spatial decorrelation than regridded MERRA\mbox{-}2. At large lags both curves approach similar sills, implying comparable large-scale variance. This matches the patterns observed during the overlap period: added fine detail without increasing the overall variation across the region.

\begin{figure}[H]
  \centering
  \includegraphics[width=0.92\textwidth]{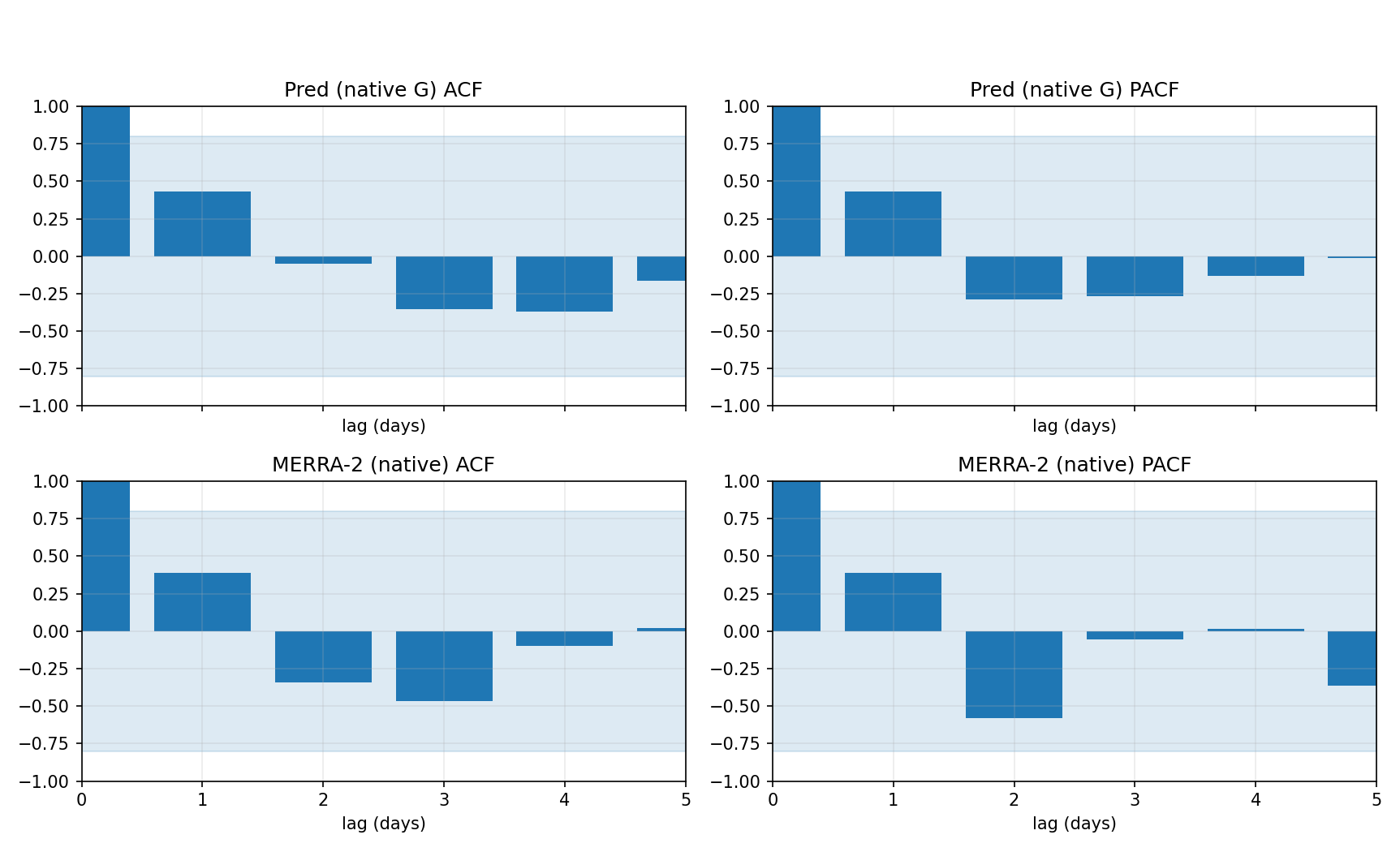}
  \caption{Temporal dependence in the OOD period (Season~3, A1, 2008). ACF and PACF of region-mean AOD for the DDPM prediction and for MERRA\mbox{-}2 regridded to the G5NR resolution. The prediction has a faster ACF decay, indicating stronger short-lag variability, while PACF mass concentrates on the first few lags in both series.}
  \label{fig:ood_acf_pacf}
\end{figure}

We next summarize temporal behavior by the daily regional mean. Figure~\ref{fig:ood_acf_pacf} reports the ACF and PACF for the prediction and for MERRA\mbox{-}2 on the same dates. The prediction retains positive short-lag correlation but its ACF decays toward zero in fewer lags than MERRA\mbox{-}2, while the PACF shows strong spikes at lags 1–2 and is near zero thereafter, indicating short daily memory. This shorter effective memory is consistent with the higher native resolution and with what was observed in the overlap period. Together with the spatial diagnostics, these results indicate that the autoregressive update maintains realistic short-lag persistence without introducing long-memory artefacts.

Across the maps, semivariograms, and ACF/PACF, the OOD results are consistent: the model follows the large-scale patterns in MERRA-2, adds sharper local detail at short distances, keeps the overall variation about the same, and shows a shorter day-to-day memory. Together with the in-data results in Section~\ref{sec: in-data downscaling result under Result section}, this indicates that the DDPM generalizes beyond its training window while preserving the G5NR-like small-scale texture learned during training.
% ============================ 4. Discussion =========================

\section{Discussion}\label{sec:Discussion}

\subsection{Model Achievements}\label{sec:Model Achievements}

Our results demonstrate that the proposed downscaling DDPM with U-Net-based transfer learning with domain similarity diagnosis and the halo-and-Hann patch stitching achieves state-of-the-art performance for satellite aerosol downscaling. This framework produces high-fidelity reconstructions that closely match the spatial semivariogram structure and temporal persistence of true G5NR, consistently outperforming baseline approaches across seasons and regions.

A notable strength of our approach is its intentionally "univariate" design, in which dust extinction is downscaled using only dust extinction without other meteorological or aerosol variables. This "univariate" formulation offers several benefits. First, it dramatically reduces data requirements, avoiding dependency on other meteorological variables that may be unavailable. Second, it preserves high domain similarity between source and target inputs, which strengthens the transfer learning robustness and reduces the risk of negative transfer. Including heterogeneous meteorological variables would introduce cross-domain shifts in scale, resolution, and spatiotemporal behavior; by contrast, focusing on a single physically homogeneous variable provides a cleaner mapping that enhances stability and robustness.

Furthermore, our results demonstrate that our proposed downscaling model achieves state-of-the-art performance for satellite aerosol downscaling, outperforming both the Artificial Neural Network Sequential Downscaling Method (ASDM) with Transfer Learning Enhancement (ASDMTE) and the Variational Downscaling Method (VDM) of \citep{Wang2022, wang_variational}, as well as recent deep learning–based downscaling models. Its superior performance arises from three technical advances: (1) robust transfer learning, (2) high-resolution generation refinement by the Diffusion Model, and (3) a physically consistent patch stitching strategy to eliminate reconstructed images' edge effect and blockiness.

VDM combines a Variational Neural Network (VNN) for the main downscaling task, ConvLSTM-based autoencoder for MERRA-2 domain pre-training, and nearest-neighbor-based stitching of training mini-batches of 15×15 image patches to reconstruct original G5NR images. Although VDM stabilizes long-term autoregressive predictions, its has several limitations. First, its transfer learning design is non-robust against negative transfer. Its source domain ConvLSTM-based autoencoder has poor learning performance of spatiotemporal dynamics within the MERRA-2 sequence, with $R^2$ far lower than that of our U-Net model trained on MERRA-2. Moreover, the VDM pipeline considers no domain divergence diagnosis but uses a naive transfer of the weak pre-trained ConvLSTM autoencoder to the downscaling task. We also found that the overall VDM performance on in-data downscaling is inferior to a VNN model trained merely on G5NR without transferring pre-trained model on MERRA-2, which suggests VDM suffers from the negative transfer. Second, the VAE-style latent sampling introduces well-known challenges, including posterior collapse, variance underestimation, and oversmoothing in reconstruction \citep{He2019LaggingIN, kingma2019introduction, Lucas2019DontBT, Bowman2015GeneratingSF}. Finally, processing only 5×5 mini-image structures restricts the model from learning regional dust transport patterns or broader-scale gradients, which are essential for realistic G5NR-scale variability. These model weakness explains why the plotted VDM fails to capture the spatial pattern in G5NR, and its predicted image pixels are uniform across the entire region.

In contrast, our downscaling DDPM avoids VAE limitations and enables recovery of sharp gradients, realistic fine-scale structure, and semivariograms whose nugget, sill, and range closely match true G5NR behavior as shown in the result section. Moreover, by employing a robust transfer learning method with domain divergence diagnosis and reusing a pre-trained U-Net with high $R^2$ on source domain for encoding G5NR sequences, the design theoretically mitigates risk of negative transfer and catastrophic forgetting and allows the downscaling model to inherit long-range spatiotemporal dependencies well learned from MERRA-2, leading to improved temporal reproduction and stable multi-day autoregressive forecasts beyond the original G5NR record. Furthermore, while the VDM and ASDMTE used a nearest-neighbor image reconstruction algorithm that causes obvious edge effects and non-smoothing reconstruction of predicted image patches into the original images, our halo-and-hann-based image stitching is specially designed to avoid reconstructing blocky and textured large images and improved both model metrics, such as $R^2$ and visual fidelity.

Collectively, these innovations yield consistently higher $R^2$, lower RMSE, coherent and stable spatial–temporal patterns, and smoother reconstructed images, compared to in-data downscaling performance in VDM and ASDMTE. Furthermore, while VDM and ASDMTE do not perform a comprehensive out-of-data downscaling evaluation, our model's predictions remain robust in out-of-data downscaling with stable temporal association, consistent lagged-error growth, and rich spatial variability comparable to the reference, which offers an additional advantage compared to VDM and ASDMTE.

Beyond VDM, the proposed framework also compares favorably with several modern deep-learning downscalers.
Deterministic U-Net–style models often produce overly smooth outputs due to their point-estimation nature \citep{Isola2016ImagetoImageTW}, whereas the diffusion model reconstructs high-frequency detail without sacrificing predictive accuracy. GAN-based directions generate sharp images but frequently suffer from model collapse and unstable training \citep{Salimans2016ImprovedTF}; DDPM avoids these issues while achieving comparable sharpness with better stability and physical consistency. Transformer-based temporal models excel at long-range temporal dependency capture but typically require very large datasets and exhibit limited ability to reconstruct fine-scale spatial detail when used alone. Finally, recent diffusion-based climate applications operate at coarser resolutions, whereas our contribution fills this gap by combining diffusion refinement with pixel-aligned satellite downscaling, a pretrained physical encoder, and a halo-weighted stitching procedure tailored to high-resolution aerosol reconstruction.

Together, these comparisons underscore that the proposed framework provides a robust, physically coherent, and computationally stable solution for multiscale satellite aerosol downscaling. It advances beyond existing variational, adversarial, diffusion-only, and transformer-based approaches.

\subsection{Limitations and future directions}\label{sec:Limitations and Future Directions}
Despite the promising in-data and out-of-data performance of our transfer–diffusion downscaling framework, several limitations remain and point to potential directions for future research.
% (1) Univariate modeling and lack of multivariate physical consistency:
Indeed, our current univariate downscaling pipeline has some advantages compared to existing downscaling models.
However, from an atmospheric-science perspective, this design introduces a fundamental limitation. The real atmospheric system is inherently multivariate: surface PM2.5 arises from a mixture of mineral dust, sea salt, black carbon, organic carbon, sulfates, and secondary production processes, many of which are represented in MERRA-2. These additional variables encode physical processes such as aerosol composition, hygroscopic growth, vertical mixing, and large-scale transport. While empirical evidence for downscaling performance gain in including additional variables is not yet conclusive, incorporating these physically relevant covariates may enable the model to learn cross-variable structure that a univariate model alone cannot capture.

Extending the current pipeline into a multivariate generative modelling could therefore potentially improve physical consistency and robustness of downscaling predictions, particularly in out-of-data downscaling. Thus, future work should explore more about multivariate generative downscaling models capable of jointly reconstructing multiple G5NR aerosol and meteorological variables. Approaches such as multitask learning \citep{9392366} or physically constrained generative models \citep{Hao2022PhysicsInformedML} could enforce cross-variable coherence, reduce uncertainty, and enable more robust and comprehensive estimation of high-resolution AOD across the entire atmospheric system.

% (2) High computational cost during inference:
Our model pipeline also demands high computational cost due to its multi-step denoising process (we specified 1000 timesteps in our DDPM) and the dense patch-based inference required for G5NR-resolution outputs. Furthermore, the autoregressive nature of the out-of-data downscaling further amplifies the computational cost of the model. Because each future day depends on the previously generated high-resolution predictions, the model must run sequentially day by day. As the prediction date moves farther away from the last available true G5NR observation, the number of required autoregressive steps grows, making each additional forecast day increasingly expensive. This sequential nature severely limits the model’s ability to perform long-horizon forecasting or to be used in operational settings. Downscaling a single day currently requires substantial GPU memory and runtime, which limits long-term forecasting and operational deployment. An important direction is to explore and find a balance between computational efficiency and prediction accuracy. One promising avenue is the adoption of latent-space diffusion architectures, such as Stable Diffusion–style models \citep{Rombach2021HighResolutionIS}, where the denoising process occurs in a compressed latent domain and decoding to pixel space occurs only once. Such designs could reduce inference time by more than an order of magnitude while retaining high-resolution detail.

%(3) Toward foundational generative models for satellite image downscaling:
While the current model pipeline is specialized in MERRA-2 to G5NR AOD downscaling, a broader need exists for general-purpose foundational models capable of downscaling diverse satellite images across multiple spatial–temporal domains. Building such a foundation would require (i) training on a wide collection of global satellite and model datasets, (ii) designing architectures that generalize across sensors, resolutions, and variables, and (iii) developing advanced transfer learning or multitask learning algorithms that allow learning in multiple domains with substantial domain divergence and rapid adaptation to unseen domains. Recent advances in spatiotemporal transformers provide a path toward such models \citep{SHU2025114872}.

% ============================ 5. Conclusion =========================
\section{Conclusion}\label{sec:Conclusion}

We presented a transfer-enhanced diffusion model for downscaling coarse satellite aerosol fields to high-resolution dust-extinction AOD. By pretraining a U-Net encoder on the full MERRA-2 record and freezing its parameters during G5NR training, the method effectively leverages rich spatiotemporal representations learned outside the limited G5NR temporal window. Domain-similarity analysis using Wasserstein distances confirms minimal divergence between MERRA-2 and G5NR, and combined with parameter freezing, this transfer-learning design mitigates the risks of negative transfer and catastrophic forgetting.

Furthermore, using a powerful diffusion model for high-resolution image generation and a halo-and-Hann patch-stitching strategy to suppress edge effects during image reconstruction, the resulting framework achieves state-of-the-art in-data downscaling performance across seasons and regions, outperforming deterministic U-Nets and VAE-based models. Spatial semivariograms and temporal autocorrelation diagnostics further show that the model preserves high-resolution spatiotemporal structure and enables stable autoregressive forecasts beyond the G5NR temporal window.

Overall, this transfer-encoder and autoregressive generative modeling framework provides a robust and physically coherent approach for producing long-term, high-resolution satellite aerosol fields. Future work could extend this research by exploring multivariate downscaling and more generalizable foundational models.

\section{Acknowledgments}\label{sec:ackowledgements}
The authors gratefully acknowledge Dr. Menglin Wang for his valuable guidance on the variational autoencoder component of this work.

% ============================ REFERENCES (placeholder) ==========================
\bibliographystyle{apalike}  
\bibliography{downscaling}

\section{Appendix}\label{sec:Appendix}
\appendix
\renewcommand{\thetable}{A\arabic{table}}
\renewcommand{\thefigure}{A\arabic{figure}}
\setcounter{table}{0}
\setcounter{figure}{0}

\FloatBarrier
\begin{table}[H]
\centering
\caption{Additional validation metrics of DDPM evaluated on held-out G5NR test days across each season and area.}
\label{tab:model-validation}

\unifiedtabsetup
{\normalsize
\setlength{\tabcolsep}{3.2pt}
\renewcommand{\arraystretch}{1.20}
\begin{adjustbox}{max width=\linewidth}
\begin{tabular}{l c c c c c c c c}
\toprule
& \multicolumn{2}{c}{\shead{Season 1}} & \multicolumn{2}{c}{\shead{Season 2}} & \multicolumn{2}{c}{\shead{Season 3}} & \multicolumn{2}{c}{\shead{Season 4}} \\
\cmidrule(lr){2-3}\cmidrule(lr){4-5}\cmidrule(lr){6-7}\cmidrule(lr){8-9}
\textbf{Model / Region} & \theadsm{A0} & \theadsm{A1} & \theadsm{A0} & \theadsm{A1} & \theadsm{A0} & \theadsm{A1} & \theadsm{A0} & \theadsm{A1} \\
\midrule
\multicolumn{9}{l}{\textbf{Large DDPM}}\\
MAE  & \cell{0.15}{0.04} & \cell{0.12}{0.02} & \cell{0.10}{0.05} & \cell{0.17}{0.04} & \cell{0.07}{0.02} & \cell{0.04}{0.01} & \cell{0.14}{0.03} & \cell{0.09}{0.02} \\
NSE   & \cell{0.83}{0.11} & \cell{0.9}{0.05} & \cell{0.85}{0.11} & \cell{0.63}{0.15} & \cell{0.94}{0.03} & \cell{0.97}{0.01} & \cell{0.78}{0.07} & \cell{0.82}{0.09} \\
KGE   & \cell{0.84}{0.06} & \cell{0.79}{0.04} & \cell{0.83}{0.09} & \cell{0.79}{0.06} & \cell{0.89}{0.04} & \cell{0.88}{0.02} & \cell{0.80}{0.05} & \cell{0.89}{0.06} \\
\bottomrule
\end{tabular}
\end{adjustbox}
}

\vspace{0.4em}
\centering
\begin{minipage}{0.9\linewidth}
\scriptsize
\emph{Notes:} Model performance is evaluated using held-out G5NR data from the test days selected for model validation. For each row, the value without the bracket is the mean of the corresponding metric computed for all test days, and the value within the bracket is the standard deviation of all computed metrics for every test days.
\end{minipage}
\end{table}
\FloatBarrier

\begin{figure}[H]
    \centering
    \includegraphics[width=0.85\textwidth]{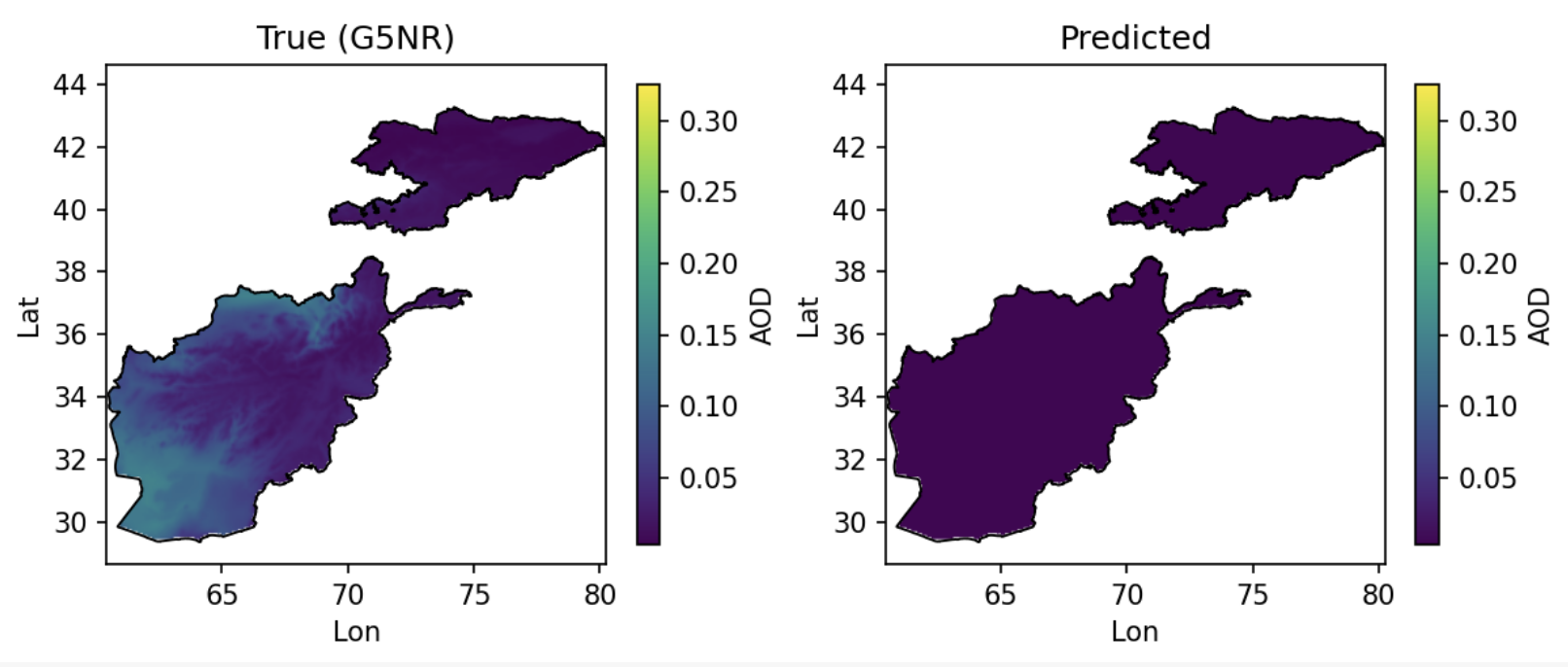} 
    \caption{Visualization of in-data downscaling prediction of VAE model (left) and true G5NR dust extinction (right) for a test day from Region 0 (Afghanistan and Kyrgyzstan) in Season 3. 
    We can see that VAE downscaled pixels are constant across the entire region and fails to capture spatial associations.}
    \label{fig:in_data_vae_prediction}
\end{figure}

% --- Appendix: extra in-data prediction visualizations ---

\begin{figure}[H]
  \centering
  % Area 0 (A0): Seasons 1–4
  \begin{subfigure}[t]{0.48\textwidth}
    \centering
    \includegraphics[width=\textwidth]{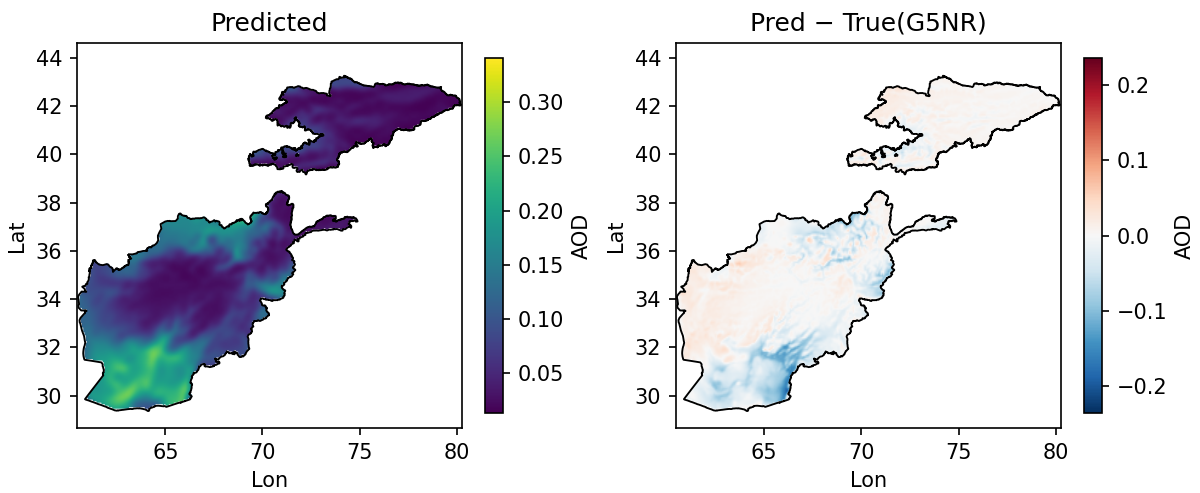}
    \caption{Season 1, A0}
  \end{subfigure}\hfill
  \begin{subfigure}[t]{0.48\textwidth}
    \centering
    \includegraphics[width=\textwidth]{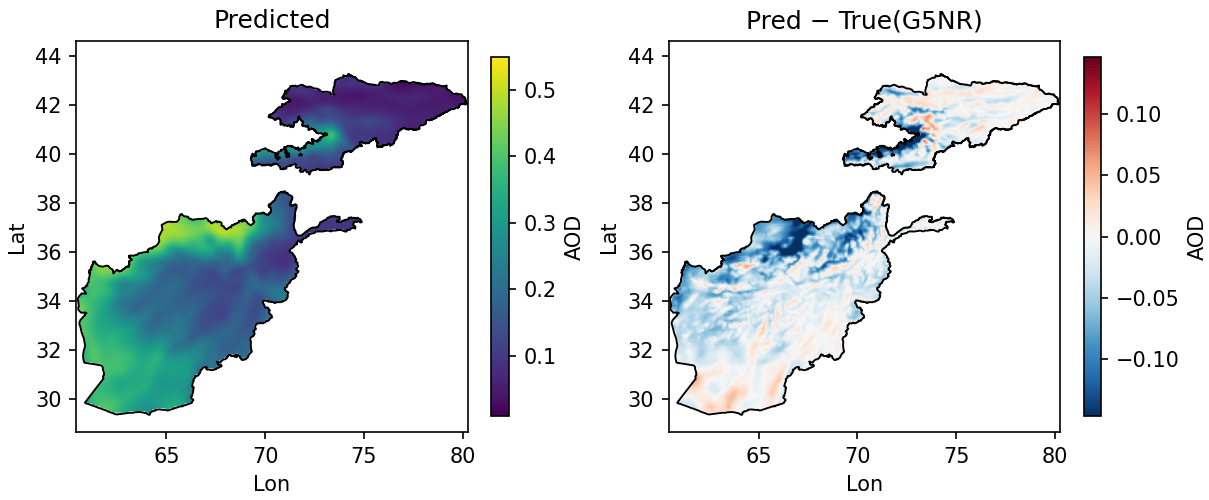}
    \caption{Season 2, A0}
  \end{subfigure}

  \vspace{0.8em}

  \begin{subfigure}[t]{0.48\textwidth}
    \centering
    \includegraphics[width=\textwidth]{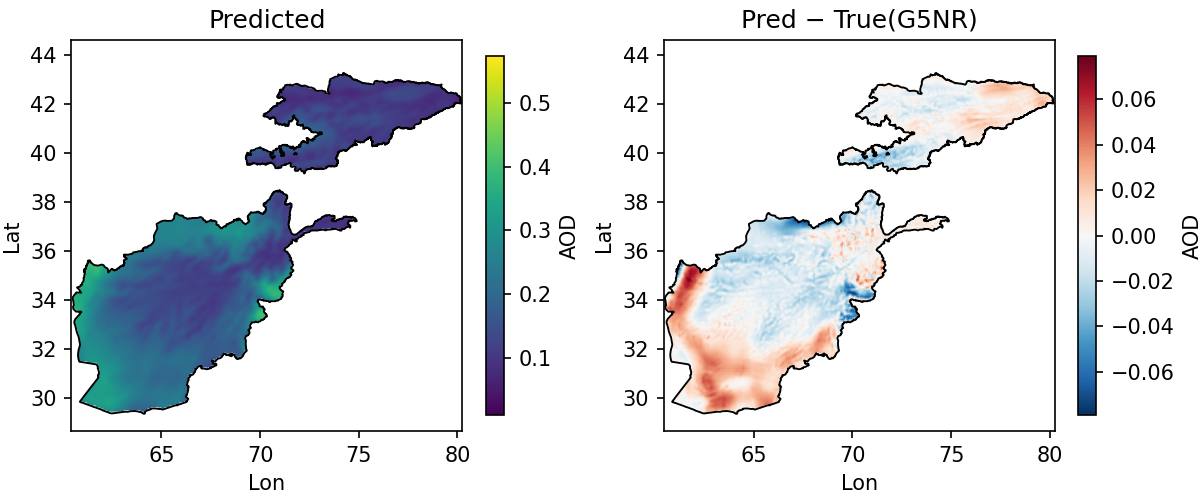}
    \caption{Season 3, A0}
  \end{subfigure}\hfill
  \begin{subfigure}[t]{0.48\textwidth}
    \centering
    \includegraphics[width=\textwidth]{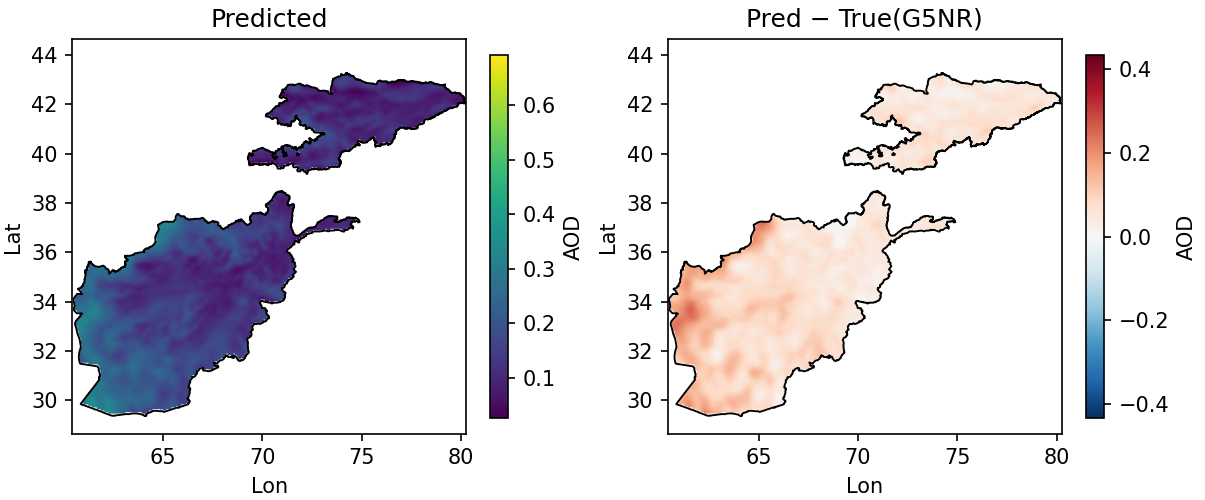}
    \caption{Season 4, A0}
  \end{subfigure}

  \caption{Additional in-data predictions of DDPM for Area 0 (A0) by each season. Each figure contains the model prediction and the pixel-wise difference, consistent with Figure~\ref{fig:in_data_prediction}.}
  \label{fig:app_in_data_A0}
\end{figure}

\end{document}